\documentclass{article} 

\usepackage{macros}
\usepackage{arxiv}

\newtoggle{arxiv}
\toggletrue{arxiv}

\usepackage{csquotes} 
\usepackage{url}            
\usepackage{graphicx}
\usepackage[numbers,sort]{natbib}
\usepackage{subfiles}
\usepackage{tabularx}
\usepackage{etoolbox}
\usepackage{multirow}
\usepackage{enumitem}
\usepackage{amssymb}
\usepackage{amsmath}
\usepackage{amsthm}
\usepackage{xspace}
\usepackage{float}
\usepackage{wrapfig}
\usepackage{booktabs}       
\usepackage{amsfonts}       
\usepackage{nicefrac}       
\usepackage{microtype}      
\usepackage{color, soul}
\usepackage{pifont}
\usepackage{nth}

\usepackage[framemethod=TikZ]{mdframed}

\usepackage{listings}
\definecolor{codegreen}{rgb}{0,0.6,0}
\definecolor{codegray}{rgb}{0.5,0.5,0.5}
\definecolor{codepurple}{rgb}{0.58,0,0.82}
\definecolor{backcolour}{rgb}{1,1,1}

\usepackage{listings}  
\usepackage{algorithm}
\algrenewcommand\algorithmicrequire{\textbf{Input:}}
\algrenewcommand\algorithmicensure{\textbf{Output:}}

\iftoggle{arxiv}{
    \setlength{\textwidth}{6.5in}
    \setlength{\textheight}{9in}
    \setlength{\oddsidemargin}{0in}
    \setlength{\evensidemargin}{0in}
    \setlength{\topmargin}{-0.5in}
    \newlength{\defbaselineskip}
    \setlength{\defbaselineskip}{\baselineskip}
    \setlength{\marginparwidth}{0.8in}
}

\newcolumntype{Y}{>{\hsize=.7\hsize}X}
\newcolumntype{Z}{>{\hsize=1.3\hsize}X}

\makeatletter
\def\@copyrightspace{\relax}
\makeatother

\makeatletter
\def\@myauthornotes{}
\def\myauthornote#1{%
  \if@ACM@anonymous\else
    \g@addto@macro\addresses{}%
    \g@addto@macro\@myauthornotes{%
      \stepcounter{footnote}\footnotetext{#1}}%
  \fi}
\makeatother

\iftoggle{arxiv}{
    \title{
        ThunderKittens: Simple, Fast, and  \textit{Adorable} AI Kernels
    }
    \usepackage{authblk}
    \author[]{Benjamin F. Spector}
    \author[]{Simran Arora}
    \author[]{Aaryan Singhal}
    \author[]{Daniel Y. Fu}
    \author[]{Christopher Ré}
    \affil[]{Stanford University}\vspace{4pt}
    \affil[]{\{\texttt{bfs,simarora,aaryan04,danfu,chrismre\}@stanford.edu}}
}

\definecolor{dkgreen}{rgb}{0,0.6,0}
\definecolor{gray}{rgb}{0.5,0.5,0.5}
\definecolor{light-gray}{gray}{0.95}
\definecolor{mauve}{rgb}{0.58,0,0.82}
\definecolor{backcolour}{rgb}{0.95,0.95,0.92}

\usepackage{graphicx}
\usepackage{subfigure}
\usepackage{wrapfig,lipsum,booktabs}
\usepackage{threeparttable}
\usepackage{makecell}
\usepackage{hyperref}
\usepackage{url}
\usepackage{algorithm}
\usepackage{algpseudocode}
\usepackage{hyperref}
\usepackage{cleveref}
\usepackage{changepage}
\usepackage{booktabs}

\usepackage{epstopdf}
\usepackage{graphicx}
\usepackage{textcomp}       
\usepackage{scalerel}       
\newcommand{\name}{\textsc{ThunderKittens}}
\newcommand{\shortname}{\textsc{TK}}

\usepackage{listings}
\usepackage{color}
\definecolor{dkgreen}{rgb}{0,0.6,0}
\definecolor{gray}{rgb}{0.5,0.5,0.5}
\definecolor{mauve}{rgb}{0.58,0,0.82}
\lstdefinelanguage{CUDACPP}{
  language=C++,
  morekeywords={__global__, __host__, __device__, __shared__, blockIdx, blockDim, threadIdx, gridDim},
  morecomment=[l][\color{magenta}]{\#},
}
\lstdefinestyle{pythonstyle}{
  language=Python,
  frame=tb,
  aboveskip=3mm,
  belowskip=3mm,
  showstringspaces=false,
  columns=flexible,
  basicstyle={\ttfamily\footnotesize},
  numbers=none,
  numberstyle=\footnotesize\color{gray},
  keywordstyle=\color[rgb]{0.13,0.29,0.53},
  commentstyle=\color[rgb]{0.13,0.55,0.13},
  stringstyle=\color[rgb]{0.31,0.60,0.02},
  breaklines=true,
  breakatwhitespace=true,
  tabsize=4
}

\usepackage{amsmath,amsfonts,bm}

\def\eqref#1{equation~\ref{#1}}

\def\1{\bm{1}}

\DeclareMathAlphabet{\mathsfit}{\encodingdefault}{\sfdefault}{m}{sl}
\SetMathAlphabet{\mathsfit}{bold}{\encodingdefault}{\sfdefault}{bx}{n}

\usepackage{enumitem}

\newcommand{\ShowNotes}{}
\newcommand{\codeword}[1]{\texttt{\textcolor{blue}{#1}}}

\ifdefined\ShowNotes
  \newcommand{\colornote}[3]{{\color{#1}\bf{#2 #3}\normalfont}}
\else
  \newcommand{\colornote}[3]{}
\fi

\definecolor{darkred}{rgb}{0.7,0.1,0.1}
\definecolor{darkgreen}{rgb}{0.1,0.5,0.1}
\definecolor{cyan}{rgb}{0.7,0.0,0.7}
\definecolor{dblue}{rgb}{0.2,0.2,0.8}
\definecolor{maroon}{rgb}{0.76,.13,.28}
\definecolor{burntorange}{rgb}{0.81,.33,0}
\definecolor{royalpurple}{rgb}{0.47,.31,0.66}

\ifdefined\ShowNotes
  
\else
  
\fi

\begin{document}

\maketitle

\begin{abstract}
    The challenge of mapping AI architectures to GPU hardware is creating a critical bottleneck in AI progress. Despite substantial efforts, hand-written custom kernels fail to meet their theoretical performance thresholds, even on well-established operations like linear attention.
    The diverse hardware capabilities of GPUs might suggest that we need a wide variety of techniques to achieve high performance. However, our work explores whether a small number of key abstractions can drastically simplify the process. We present \name{} (\shortname{}), a framework for writing performant AI kernels while remaining easy to use and maintain. 
    Our abstractions map to the three levels of the GPU hierarchy: (1) at the warp-level, we provide 16x16 matrix tiles as basic data structures and PyTorch-like parallel compute operations over tiles, (2) at the thread-block level, we provide a template for overlapping asynchronous operations across parallel warps, and (3) at the grid-level, we provide support to help hide the block launch and tear-down, and memory costs. We show the value of \shortname{} by providing kernels that match or outperform prior kernels for a range of AI operations. We match CuBLAS and FlashAttention-3 on GEMM and attention inference performance and outperform the strongest baselines by $10-40\%$ on attention backwards, $8\times$ on state space models, and $14\times$ on linear attention.
\end{abstract}

\vspace{-4mm}
\section{Introduction}
\vspace{-1mm}
AI is bottlenecked by the problem of efficiently mapping AI architectures onto accelerated GPU hardware.
There has been a Cambrian explosion of ML architectures \citep{ho2020denoising, gu2023mamba}; however, the performance of these architectures remains substantially below their theoretical potential, despite substantial effort to develop \textit{kernels}, or GPU implementations.
Notably, kernel support has been poor even for softmax attention, which is used throughout industry. FlashAttention-2 \citep{dao2023flashattention2} suffered a 47\% performance degradation when translated to the H100 GPU, and it took over two years from the release of the H100 to develop FlashAttention-3 \citep{dao2024flashattention3}.

We are inspired by several approaches to supporting the development of AI kernels. 
Ideally, we would have a framework that supports high \textbf{performance} for a \textbf{breadth} of primitives, while being \textbf{easy to use}, learn from, and maintain.
High performance C++ embedded libraries like NVIDIA CUTLASS/CuTe~\citep{nvidia2017cutlass} contain a myriad of nested templates, while compiler based approaches like Triton~\citep{triton} provide users with simpler interfaces, but fewer optimizations. We ask how broad and fast we can go by choosing a small and opinionated set of abstractions. 

The main vector of growth for accelerated compute is in specialized matrix multiply units. On the NVIDIA A100 and NVIDIA H100 GPUs, BF16 tensor cores represent $16\times$ the FLOPs available relative to general-purpose BF16 / FP32 compute. Consequently, any high performance framework must prioritize keeping tensor cores at high utilization whenever possible. However, all kernels have other operations too (like memory loads or the softmax in attention), and it is crucial to minimize the overhead of
non-tensor core operations. This proposition is at the heart of our approach.

\begin{figure}
    \centering
    \includegraphics[width=\textwidth]{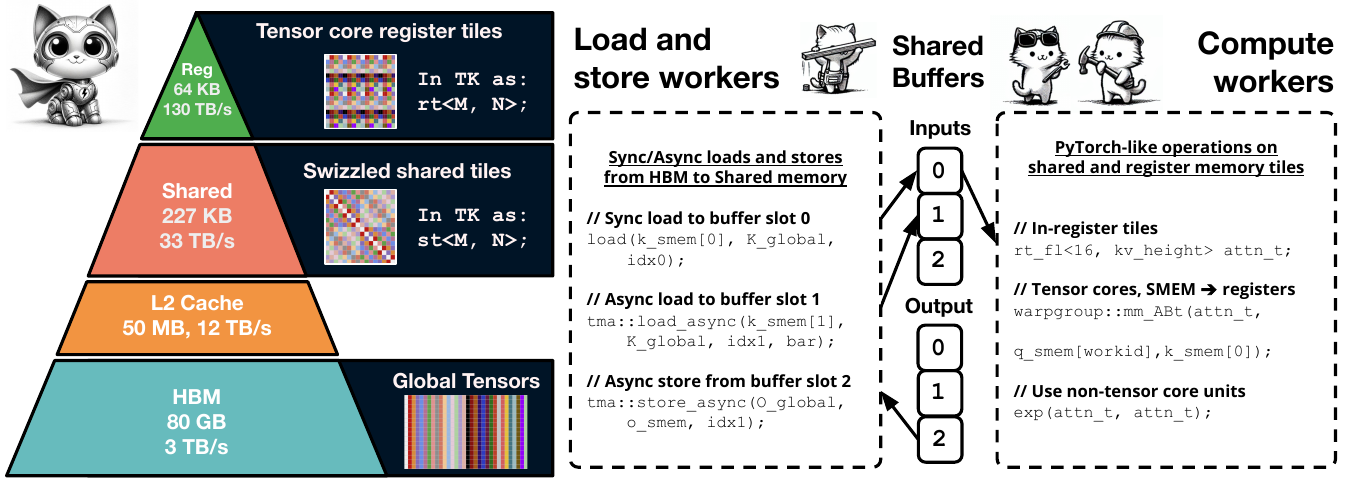}
    \vspace{-1.5em}
    \caption{\name{} explores whether a small set of abstractions can support efficient and simple AI kernels. Inspired by PyTorch, our abstractions include tiles with managed layouts and operations over tiles. We provide a general program template for coordinating asynchronous parallel workers -- e.g., workers that load from and store data to HBM, while other workers perform computation in fast memory.
    }
    \label{fig:main_fig}
    \vspace{-4mm}
\end{figure}

To understand the complexities and opportunities in building a simple, yet high performance framework, we examine a simplified model of GPU parallelism, further detailed in section 2.\footnote{We discuss NVIDIA, but the parallelism types hold across architectures, including AMD and Apple GPUs.}
\begin{enumerate}[itemsep=0.1pt,topsep=0pt,leftmargin=*]
    \item \textbf{Warp-level parallelism:} 
    Modern GPUs consist of tens of thousands of hardware threads which execute in parallel. Threads are organized into small groups, ``warps'', which execute instructions together. Memory \textit{layouts} determine how the logical data elements are mapped to physical thread ownership. If multiple threads try to access the same region (``bank'') of memory, this can create expensive serializations between the threads (called ``bank conflicts'').
    
    \item \textbf{Block-level parallelism:} Warps are grouped into ``blocks'' of threads, which can quickly share data. Warps execute their instructions on physical execution units, and having more warps in a block (called \textit{occupancy}) can help run more instructions at the same time, reducing runtime. For example, one warp can run tensor cores for matmul, while another uses the ALU for $\max$.
    \item \textbf{Grid-level parallelism.} GPUs run many blocks of threads at once, which communicate through large but slow global memory (HBM). An on-chip shared L2 cache helps reduce memory latencies and increase bandwidth if thread blocks reuse the same data. Thread blocks also face setup and tear-down latency overheads, called ``pipeline bubbles''
\end{enumerate}
\vspace{1mm} 

Despite the apparent need for a myriad of techniques to leverage all these  hardware capabilities, our central technical finding is that indeed, for many AI kernels, \textit{a small number of key abstractions exist that can simplify the process of writing high-performance kernels}.
Our exploration led us to develop \name{} (\shortname{}), an AI kernel framework built around three key principles:
\vspace{1mm} 
\begin{enumerate}[itemsep=0.1pt,topsep=0pt,leftmargin=*]
    \item \textbf{Tile data structures with managed layouts:} Our interface is inspired by familiar ML frameworks like PyTorch and NumPy~\citep{paszke2019pytorchimperativestylehighperformance}, as highlighted in \Cref{fig:listing_1}. At the warp level, we use a $16 \times 16$ matrix tile as our basic data structure, maximizing compatibility with and encouraging the use of tensor cores. 
    \shortname{} automatically picks the optimal memory layouts for the tiles to minimize bank conflicts while remaining compatible with specialized hardware instructions, avoiding user effort. 
    We provide a set of parallel compute primitives over tiles, based on the suite of operations in PyTorch (e.g., pointwise $\mathrm{multiply}$, $\mathrm{mma}$, $\exp$, and $\mathrm{cumsum}$ over tiles).
    \item \textbf{Program template for asynchronous work:} At the block level, \shortname{} provides a general kernel template for coordinating asynchronous execution across warps in a thread block, built on the producer-consumer paradigm~\citep{dijkstraPC}.
    The developer's effort reduces to populating a few  boilerplate functions within this model, using our PyTorch-like operands, and the template internally hides latencies through memory pipelines and synchronization primitives  (\Cref{fig:main_fig}). 
    \item \textbf{Grid scheduling for pipelining thread-blocks.} At the grid level, we show
    \shortname{} can help developers reduce pipeline bubbles and improve L2 cache hit rates. 
    Our template supports a \textit{persistent grid}, where we overlap
     memory loads across thread block boundaries.
\end{enumerate}

\clearpage
\vspace{1mm} 
\noindent We highlight the value of these abstractions for developers in two ways:
\begin{itemize}[itemsep=0.1pt,topsep=0pt,leftmargin=*]
    \item Through our exploration, we identify a few fundamental tradeoffs between achieving different types of parallelism including in setting the tile layouts (warp-level), occupancy (block level), and block launch order (grid level). Through our ablation studies (\Cref{sec3_method}), we show how the simplified interface in \shortname{} gives users the control to navigate the tradeoffs.
    \item We validate the \shortname{} abstractions by providing kernels that match or outperform prior kernels for a range of AI operations. We match CuBLAS GEMMs and FlashAttention-3 attention inference, and outperform the strongest baselines by $10-40\%$ on attention backwards, up to $8\times$ on state space models, and up to $14\times$ on linear attention.
    These kernels are written by a small academic team, including by undergraduates with no prior CUDA experience.
\end{itemize}

\begin{figure}[t]
\begin{minipage}{0.39\textwidth}
\begin{adjustwidth}{2em}{-2.5em} 
PyTorch attention:
{\Huge
\begin{lstlisting}
# imports 
import torch 
import torch.nn.functional as F


# compute Q@K.T
att = torch.matmul(
    q, k.transpose(2, 3)) 
    
# compute softmax
att = F.softmax(
    att, dim=-1, 
    dtype=torch.float32)

# convert back to bf16
att = att.to(q.dtype)

# mma att@V
attn_output = torch.matmul(att, v)
\end{lstlisting}
}
\end{adjustwidth}
\end{minipage}  
\hfill
\begin{minipage}{0.45\textwidth}
\begin{adjustwidth}{0em}{-2em} 
\name{} attention:
{\Huge
\begin{lstlisting}
// imports
using namespace kittens;
rt_bf<16, 64> k_reg, v_reg;
// load k from shared memory to register
load(k_reg, k_smem[subtile]); 
// compute Q@K.T
zero(att);
mma_ABt(att, q_reg, k_reg, att); 
// compute softmax
sub_row(att, att, max_vec); 
exp(att, att); 
div_row(att, att, norm_vec);
// convert to bf16 for mma_AB
copy(att_mma, att); 
// load v from shared memory to register 
load(v_reg, v_smem[subtile]); 
auto &v_reg_col = swap_layout_inplace(v_reg); 
// mma att@V onto o_reg 
mma_AB(o_reg, att_mma, v_reg_col, o_reg); 
\end{lstlisting}
}
\end{adjustwidth}
\end{minipage}
\vspace{-2mm}
\caption{Attention implemented in \shortname{} using familiar PyTorch-like operations on tiles.}
\vspace{-4mm}
\label{fig:listing_1}
\end{figure}

Our contributions are (1) showing a small and opinionated set of abstractions in \shortname{} that goes surprisingly far for writing simple and performant kernels; and (2) providing a collection of performant AI kernels.
\shortname{} kernels are in production at ML inference providers and high-frequency trading firms alike. 
We hope that \shortname{} and its insights help improve the accessibility of AI kernels.

\vspace{-1mm}
\section{GPU fundamentals}
\vspace{-2mm}
\label{sec2_preliminaries}

\begin{wrapfigure}{rt}{0.35\linewidth}
\vspace{-0.7cm}
\centering
\includegraphics[width=\linewidth]{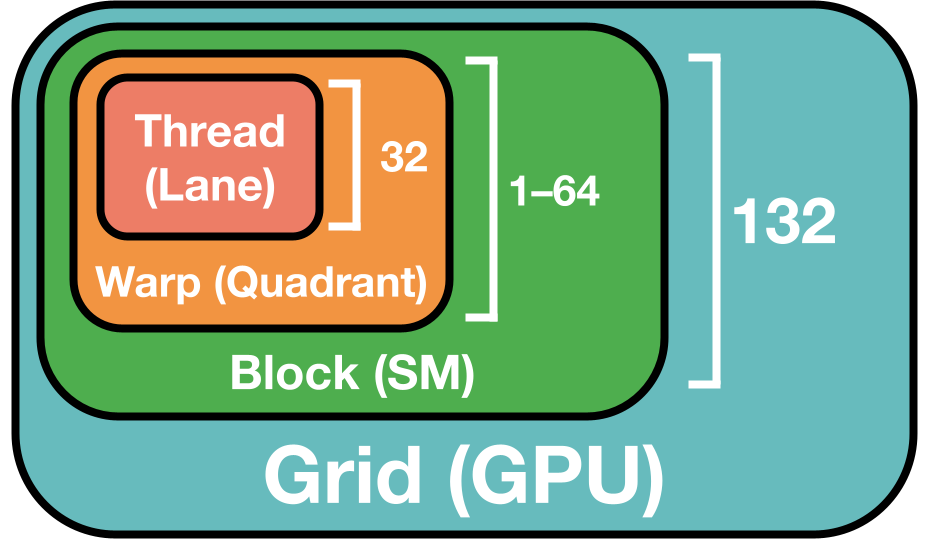}
\vspace{-0.6cm}
\caption{The software (and \\physical) GPU hierarchy.}
\vspace{-0.6cm}
\label{fig:gpu_graphic}
\vspace{-2mm}
\end{wrapfigure}

GPU tasks are divided into small programs called \textit{kernels}. A kernel loads data from high bandwidth memory (HBM), performs work on it, and writes the outputs back to HBM before concluding. Before we explain \name{}'s abstractions, we provide background on GPU parallelism including \textit{warp}, \textit{block} and \textit{grid-level} parallelism. We follow NVIDIA's terminology and focus on the H100 SXM GPU, though the principles apply across GPU vendors and generations.

\vspace{-1mm}
\subsection{GPU hierarchy} The GPU software hierarchy closely follows its physical hardware hierarchy (\Cref{fig:gpu_graphic}). Here, we illustrate several of its most important components and aspects.

\begin{enumerate}[itemsep=0.1pt,topsep=0pt,leftmargin=*]
    \item \textbf{Warps} consist of groups of $32$ nearby \textit{threads} that operate on data in small but fast \textit{register} memory. These instructions run on various physical \textit{execution units}, which are specialized for different compute operations (below) and different threads can occupy different units simultaneously:
    \begin{enumerate}
        \item Load and store units, to bring data into and out of registers. Advanced GPUs have also introduced dedicated hardware acceleration for bulk loads and stores.
        \item General purpose compute pipelines, such as \textbf{ALU} for $\max, \min$, \textbf{FMA} for multiplies and adds, and \textbf{XU} for complex operations like $\exp$. Throughput differs across the pipelines.
        \item Accelerated matrix multiply hardware (tensor cores), which have most of the GPU compute.
    \end{enumerate}
    
    \item \textbf{Thread blocks} are groups of warps which together execute a kernel on a physical core, called a \textit{streaming multiprocessor} (SM). Although each SM has just four physical execution units, up to 64 software warps can simultaneously run on it (called ``occupancy''). These collocated warps often contend on hardware resources: registers, shared memory, issue slots, and compute pipelines, but together they can help keep many work streams running at the same time within each execution unit. Warps \textit{synchronize} at barriers, during which they cannot issue new work.

    Importantly, warps within the same block can quickly communicate through special \textit{shared memory} (SMEM, 227 KB, 33 TB/s). To improve bandwidth, SMEM is grouped into 32 physical ``banks'' of memory, which can serve memory simultaneously. However, if different threads try to access the same bank at the same time (called a \textit{bank conflict}), their accesses must be serialized, which both increases access latencies and reduces available bandwidth. 
    Hopper has limit of 255 registers per thread and attempts to request more, results in \textit{spills} to the L1 cache. SMEM can be reallocated as an \textit{L1 cache} for fast access to frequently used memory like spilled registers.

    \item \textbf{Grids} of multiple thread blocks are launched to run the kernel. The H100 SXM GPU has 132 physical SM's which can run thread blocks at the same time. Although SM's are capable of collocating multiple thread blocks, most AI kernels can achieve high performance by simply collocating more warps within a single thread block (increasing the occupancy).

    Thread blocks on the same GPU share common memory resources: large but slow high-bandwidth memory (80 GB, 3 TB/s), which has both the greatest latency and least bandwidth of all GPU memory, and a smaller but faster L2 cache (50 MB, 12 TB/s).

    There are overheads to scheduling blocks. First, the block launch incurs \textit{setup} costs and although this cost must be paid at least once at the initial kernel launch, kernels that continuously launch many large blocks can incur further costs. Second, there are \textit{tail effect} costs if the grid is sized poorly.
    If a kernel of 133 blocks is executed on an H100 with 132 physical SMs, the kernel would require two waves to execute, the first with full efficiency, and the second with $<1\%$ efficiency.
\end{enumerate}

\vspace{-1mm}
\subsection{Cost model}
\vspace{-2mm}
\label{eqn:model2cost}
Beyond reducing the total amount of work, the other key way to reduce execution time is to overlap multiple kinds of work at once. Summarizing the above components, we provide a simplified cost model for GPU parallelism. We break down the overall kernel execution time $\mathbf{C}_{\mathrm{Overall}}$ as:
\vspace{-1mm}
\begin{align}
    \mathbf{C}_{\text{Overall}} = 
    \max\Big(
    \underbrace{
    \mathbf{C}_{\text{HBM}}, 
    \mathbf{C}_{\text{L2}}, 
    \mathbf{C}_{\text{L1}}, 
    \mathbf{C}_{\text{Shared}}
    }_{\textbf{Memory}},
    \underbrace{
    \mathbf{C}_{\text{Tensor}},
    \mathbf{C}_{\text{ALU}},
    \mathbf{C}_{\text{FMA}},
    \mathbf{C}_{\text{XU}}
    }_{\textbf{Compute}}
    \Big) +
    \underbrace{
    \mathbf{C}_{\text{Setup}} + 
    \mathbf{C}_{\text{Sync}}
    }_{\textbf{Overhead}}
    \nonumber
\end{align}
where memory costs are a combination of the latency and bandwidth, and compute costs are a combination of latency and throughput.

This model represents the \textit{ideal case} of perfect overlapping between memory, compute, and tensor core costs. A kernel's actual performance will lie between the $\max$ and the $\mathrm{sum}$ of these components, depending on the workload properties (\textit{i.e.}, some operations are inherently sequential), as well as the efficiency of its implementation. Nonetheless, our explorations will be guided by trying to (1) reduce these individual costs, and (2) improve their collective overlapping.

\subsection{GPU programming frameworks}
We are inspired by a number of related efforts to simplify the development of AI kernels, such as NVIDIA CUTLASS/CuTe ~\citep{nvidia2017cutlass} and Triton~\citep{triton}.

CUTLASS's myriad of nested CUDA templates helps power highly optimized AI kernels~\citep{dao2024flashattention3, bikshandi2023acase, bikshandi2023developing} and fundamentally, the same kernels are expressible in \shortname{} and CUTLASS, since both are \textit{embedded} libraries, giving users the full power of C++. We take a complementary approach by being rather opinionated about the abstractions. We ask: \textit{(1) How far can we get with a small of templates? and (2) Does concision sacrifice performance?} An appealing outcome is improved accessibility to AI researchers, since it can be challenging to fully leverage the capabilities of CUTLASS~\citep{bikshandi2023developing}. We find that even industrially popular kernels written in CUTLASS, like FlashAttention-3, struggle from  preventable issues like bank conflicts. We seek abstractions that manage such issues for users. Most recent AI architectures use high level compilers instead~\citep{dao2024transformers, yang2024fla, fu2023flashfftconv}. 

Triton, PyTorch~\citep{paszke2019pytorchimperativestylehighperformance}, TVM~\citep{tvm}, TensorFlow XLA~\citep{tensorflow}, and others approach the problem from a compiler perspective. 
The frameworks are not C++ embedded, so it can be challenging to use unsupported specialized hardware instructions. It can also be difficult to manage asynchronous execution and register usage in high level frameworks.
We explore avenues that retain the simple, PyTorch-like feel \textit{and} enable high performance in the next section. An extended discussion of related work is in \Cref{app:related_work}.

\vspace{-2mm}
\section{ThunderKittens}
\vspace{-1mm}
\label{sec3_method}
\vspace{-1mm}
We present \name{} (TK), a framework designed to simplify the development of high-performance AI kernels while leveraging the full capabilities of modern GPUs. 
This section (1) introduces our key programming abstractions and (2) shows how they can help developers navigate the tradeoffs between different types of parallelism. \Cref{sec3_model_1} focuses on warp level, \Cref{sec3_model_2} on thread block level, and \Cref{sec3:sec3_grid} on grid level parallelism. 

As running examples in this section, we show how \shortname{} helps optimize attention~\citep{vaswani2018attention} and GEMM kernels. \Cref{sec:sec4_results} demonstrates how the principles yield performant kernels for a breadth of AI operations (\textit{e.g.}, attention variants, convolution, SSM, rotary).

\vspace{-2mm}
\subsection{Warp parallelism with familiar data structures and operations}
\label{sec3_model_1}
\vspace{-1mm}

At its core, \name{} is built on two fundamental abstractions -- \textbf{tile data structures} at each level of the memory hierarchy and \textbf{bulk operands on tiles} akin to the familiar suite of operations in PyTorch and NumPy. We first define the abstractions, and then show they can help developers navigate tradeoffs between the tile \textit{sizes} and efficiency.

\vspace{-2mm}
\paragraph{Programming abstractions}
\shortname{} is heavily inspired by PyTorch and NumPy, given their familiarity to ML audiences~\citep{paszke2019pytorchimperativestylehighperformance}.
We provide a concise set of parallel compute operations, based on the suite of operations in PyTorch (e.g., in \Cref{fig:listing_1}). The operations are executed by a ``worker'' abstraction, or a warp or warpgroup ($4$ warps) of threads that collaboratively own and operate on a piece of data. 
\shortname{} uses a $16 \times 16$ matrix tile as its basic data structure, designed to maximize compatibility with tensor cores. We provide tiles for each level of the memory hierarchy:
\begin{enumerate}[itemsep=0.1pt,topsep=0pt,leftmargin=*]
    \item Register tiles and vectors, which are templated by type, shape, and layout. In \Cref{fig:listing_1} we initialize a bfloat16 type tile with a column-major layout, height $16$, width $64$.
    \item Shared tiles and vectors, which are templated by type and shape. 
    \item Global layout descriptors: We set up HBM loads and stores as indexing into $4$D tensors (similar to $\{$\texttt{batch}, \texttt{head}, \texttt{length}, and \texttt{embed}$\}$ in PyTorch). Dimensions can be known at compile-time or runtime. Compile-time dimensions can be stored in the instruction cache, saving registers.
\end{enumerate}

An advantage of these tile-based abstractions is that they enable \shortname{} to statically check layouts and operations, which is important because GPU kernels are often difficult to debug. For example, an in-register tensor core multiply \texttt{mma\_AB} requires $A$ to be in a row-major layout, and $B$ to be in a column-major layout, and \shortname{} can raise compile-time errors if these conditions are not met. 

\begin{figure}[h!]
    \centering
    \vspace{0.125cm}
    \noindent\makebox[\linewidth]{
        \includegraphics[width=\linewidth]{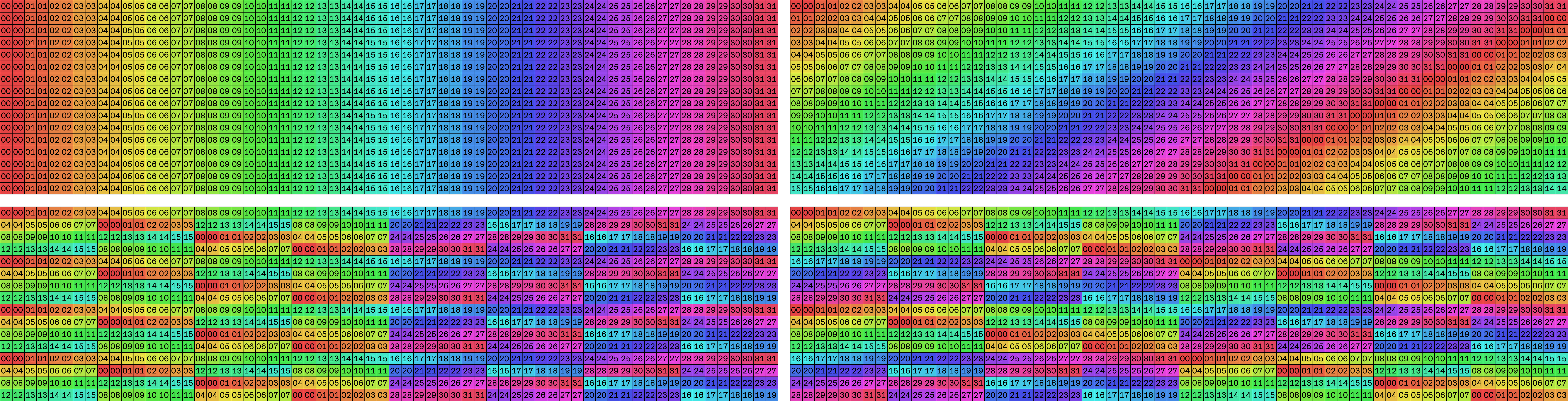}
    }
    \vspace{-0.5cm}
    \caption{Shared memory bank layouts, illustrated for a 16x64 16-bit tile. \textbf{Top left:} A naive, row-major layout. Although loading rows is efficient, loading into a tensor core layout suffers 8-way bank conflicts. \textbf{Top right:} A padded layout, which has no bank conflicts but consumes additional memory and has poor hardware support. \textbf{Bottom:} Two of \shortname{}'s three chosen layouts, with compile-time selection based on width. (Bank conflicts are unavoidable for some tile sizes while maintaining good hardware support.) These layouts have 2-way and no bank conflicts, respectively.}
    \label{fig:shared_layouts}
\end{figure}

\vspace{-1mm}
\paragraph{Choosing a memory layout} Layouts specify how logical data elements are mapped to physical thread ownership. Different execution units, tile sizes and types, and hardware-accelerated instructions benefit from different layouts. A poor choice of layout can lead to bank conflicts ($\mathbf{C}_{\text{Shared}}$, \Cref{sec2_preliminaries}). 
Our goals are:
\begin{itemize}[itemsep=0.1pt,topsep=0pt,leftmargin=*]
    \item We want our register tiles (the fastest GPU memory) to by-default keep memory in the layouts required by tensor core units (the fastest GPU compute units). Shown in \Cref{fig:main_fig} (Left), where each color represents a different thread's ownership over the data elements, the tensor formats are rather difficult to use and reason about, highlighted in our discussion of naive layouts in \Cref{fig:shared_layouts}.
    \item We want to support the use of hardware-accelerated instructions (\textit{e.g.}, asynchronous matrix multiply and bulk copy instructions), which also require specific shared memory layouts. 
\end{itemize}

In \shortname{}, we simplify the search space to $3$ layouts -- strided at 32, 64, and 128 byte intervals -- and automatically give shared tiles with the largest layout that the tile size supports to minimize bank conflicts. 
Highlighted in \Cref{sec4:exotic_kernels}, even optimized FlashAttention-3 kernel written using CUTLASS and CuTe templates suffer from bank conflicts, hurting performance. Our approach helps minimize conflicts.

\vspace{-2mm}
\subsection{Block parallelism with a generalized asynchronous template}
\label{sec3_model_2}
\name{} helps developers reduce overheads by coordinating how workers in a thread block asynchronously overlap execution.
 Though the GPU hierarchy might suggest that we need a wide variety of techniques, we propose a \textit{single} concise template that we find enables high performance on a surprisingly broad range of AI workloads.
We first define the template, which has four steps -- load-compute-store-finish (LCSF for short) -- and builds on the classical producer-consumer paradigm~\citep{dijkstraPC, bauer2011cudadma}. We then show how the LCSF template can help developers navigate the tradeoffs between occupancy and efficiency.

\begin{figure}[h]
  \begin{minipage}{0.35\textwidth}
Load function:
\begin{lstlisting}
if(warpgroup::warpid() == 0) {
    tma::expect(inputs_arrived,
        block.k, block.v);
    tma::load_async(
        block.k, globals.k,
        {batch, head, iter, 0},
        inputs_arrived);
    tma::load_async(
        block.v, globals.v,
        {batch, head, iter, 0},
        inputs_arrived);
}
else arrive(inputs_arrived);
\end{lstlisting}  \end{minipage}
  \hfill
  \begin{minipage}{0.58\textwidth}
Compute function:
\begin{lstlisting}
warpgroup::mm_ABt(att, scratch.q[state.id], block.k);
warpgroup::mma_async_wait();

// softmax (simplified)
sub_row(att, att, max_vec);
exp(att, att);
div_row(att, att, norm_vec);

copy(att_bf16, att); 

warpgroup::mma_AB(state.o, att_bf16, block.v);
warpgroup::mma_async_wait();
arrive(inputs_finished);
\end{lstlisting}
  \end{minipage}
\vspace{-2mm}
\caption{An attention kernel within the LCSF template. The left shows the functionality for workers focused on managing HBM to SRAM memory movement, and the right shows the functionality for parallel compute workers that operate in fast memory, registers and SRAM. 
}
\label{fig:pc_attn_listing}
\end{figure}

\vspace{-2mm}
\paragraph{Programming abstractions} As discussed in ~\Cref{sec2_preliminaries}, the general pattern of an AI kernel is to load tiles of large tensors from HBM to SRAM, perform computation in fast memory, store the result for the tile back to HBM, and repeat this for the next tiles. To use the LCSF template, the developer writes four functions:
\vspace{2mm}
\begin{enumerate}[itemsep=0.1pt,topsep=0pt,leftmargin=*]
    \item Load function: The load function specifies the data that load workers should load from HBM to shared memory, and when to signal to compute workers that this memory is ready for use. 
    \item Compute function: This function specifies the kernel instructions that compute workers should execute, using the tile data structure and operation primitives from \Cref{sec3_model_1}.
    \item Store function: The store function specifies what data needs to be stored to HBM by store workers.
    \item Finish function: At the end of the kernel, the workers store any final state and exit.
\end{enumerate}
\vspace{2mm}

\noindent \shortname{} provides abstractions to help the developer manage worker overlapping and synchronization.

\begin{wraptable}{r}{0.4\textwidth}
\begin{center}
\vspace{-5mm}
\centering
\begin{tabular}{ccc}
\toprule
M = N = K & Stages & TFLOPS \\ 
\midrule
4096 & 1 & 260 \\ 
4096 & 2 & 484 \\ 
4096 & 3 & 683 \\ 
4096 & 4 & 760 \\ 
\bottomrule
\end{tabular}
\label{tab:pc_stages}
\caption{\textbf{Pipeline buffer stages} We measure efficiency in TFLOPS for our GEMM kernels as we vary the number of pipeline buffer stages in the \shortname{} template.}
\vspace{-9mm}
\end{center}
\end{wraptable}
1. Multi-stage buffer: The template maintains $N$-stage \textit{pipelined buffers} in shared memory, which are used for loads and stores from HBM. Load/store workers add/remove tiles of data from the buffers, based on the status of compute workers.  With a single stage, load workers would need to wait for all compute workers to finish executing before replacing the input tile. A 2-stage buffer can hide the HBM load (store) latency since the next tile can asynchronously load, while the compute workers execute on the current tile. Deep buffers can reduce the amount of synchronization required across compute workers, allowing them to operate on different tiles concurrently. 

\shortname{} lets the user set a single number to specify the number of stages, and manages the setup and use of these buffers for the user. We demonstrate this in \Cref{tab:pc_stages}, where we vary the number of stages $N \in \{1, 2, 3, 4\}$ for our GEMM kernel.
\newline

2. Synchronization barriers: Load/store workers need to alert compute workers when new memory is written to the input buffer. Compute workers need to alert load/store workers when tiles are written to the output buffer, or when input tiles can be evicted from the input buffer. Within the \shortname{} template, we provide an \texttt{arrive} function for workers to signal that they have finished their stage. \newline

3. Asynchronous I/O: We wrap synchronous and asynchronous load and store instructions, including \texttt{cp.async} and TMA, in the same interface. We automate tensor map descriptor creation for TMA hardware-accelerated address generation for our global layout descriptors (\texttt{gl}).

\vspace{-2mm}
\paragraph{Tradeoffs between occupancy and efficiency}

\begin{wrapfigure}{rt}{0.5\linewidth}
    \centering
    \vspace{-5mm}
    \includegraphics[width=0.5\textwidth]{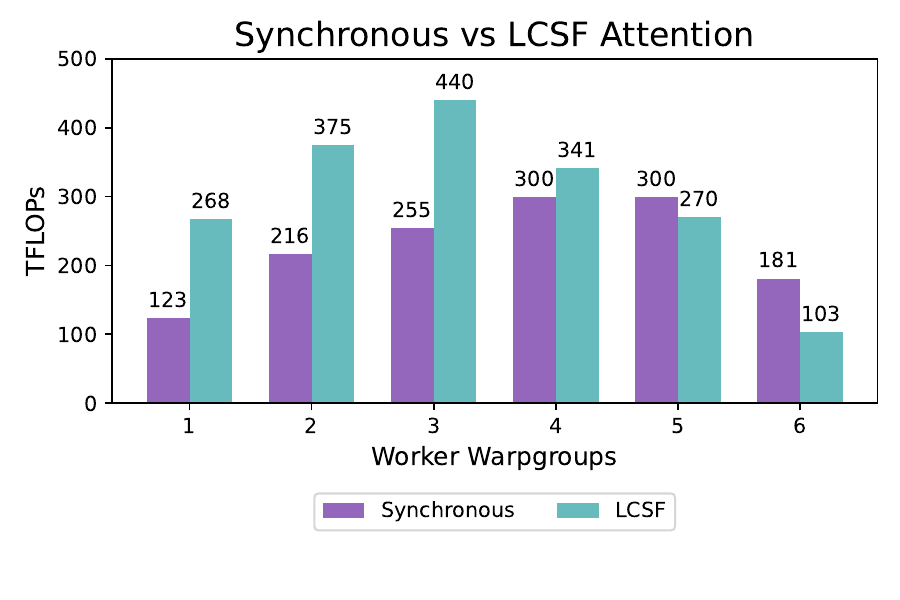}
    \vspace{-3em}
    \caption{\textbf{Occupancy tradeoff:} \textbf{(Left)} Attention TFLOPs as a function of occupancy, benchmarked with head dimension 64 and a sequence length of 4096. We compare a basic synchronous and LCSF kernel.
    }
    \label{fig:gantt}
    \vspace{-6mm}
\end{wrapfigure}
\shortname{} parametrizes  the \textit{number} of load/store and compute workers (or occupancy)  
providing a simple way for developers tune their kernels.
As discussed in \Cref{sec2_preliminaries}, higher occupancy increases overlapping, but creates contention over limited hardware resources (e.g., registers). With fewer registers, workers need to operate on smaller tiles of data, resulting in more instruction issues, SRAM to register I/O, and potentially higher synchronization costs due to the increased data partitioning across workers. 
 
\Cref{fig:gantt} shows the occupancy tradeoffs for attention kernels. 
We consider (1) a simple kernel that only uses warp level parallelism (Listing \ref{fig:listing_1}) and (2) a kernel written in the LCSF template (Listing \ref{fig:pc_attn_listing}).
With \textit{both} kernels, performance increases until resource contention dominates.
Second, we observe that \textbf{LCSF expands the Pareto frontier} beyond the warp-level parallel kernel as we vary occupancy. \newline

\noindent We find the general LCSF template to be effective across a range of AI workloads. We keep the template lightweight and simple by making opinionated design choices. However, we don't want \shortname{} to get in the way of achieving peak GPU performance -- in contrast to frameworks like Triton, \shortname{} is \textit{embedded}, meaning developers can use the full power of C++ to extend the library as warranted.

\vspace{-2mm}
\subsection{Grid parallelism with block launch scheduling}
\label{sec3:sec3_grid}
Next, at the grid level, we explore how to coordinate thread block launches. \shortname{}'s template does not explicitly choose grid structures for the user, however we provide a tradeoffs study of two key opportunities: reducing the setup and tear-down costs for each thread block ($\mathbf{C}_{\text{Setup}})$, and encouraging memory reuse between thread blocks to avoid slow HBM accesses ($\mathbf{C}_{\text{HBM}})$. 

\begin{wraptable}{r}{0.5\textwidth}
\vspace{-3mm}
\centering
\begin{tabular}{lcccc}
\toprule
M=N &  K  & \shortname{}-No & \shortname{}-Yes & CuBLAS \\ 
\midrule
4096 & 64   & 93  & 108 & 69  \\ 
4096 & 128  & 161 & 184 & 133 \\ 
4096 & 256  & 271 & 309 & 242 \\ 
4096 & 512  & 414 & 450 & 407 \\ 
4096 & 1024 & 565 & 600 & 633 \\ 
\bottomrule
\end{tabular}
\vspace{-2mm}
\caption{\textbf{Persistent block launch} TFLOPS for \shortname{} GEMM kernels with (\textbf{yes}) persistent and without (\textbf{no}) persistent launch as we vary matrix dimension $K$.}
\vspace{-6mm}
\label{tab:persistent_table}
\end{wraptable}
\paragraph{Block launch costs} We first explore the use of a \textit{persistent grid}, where we launch thread blocks on the full set of 132 SMs upfront, and simply load the next chunk of work for the kernel within the existing block, instead of launching a new block. We also explore the idea of having the block load shared memory into the input stage of our template's memory buffer to prepare for the next chunk of work, while the thread block runs the finish stage for the prior chunk of work. \Cref{tab:persistent_table} shows the benefit of the optimizations for our GEMM kernels.

\paragraph{L2 reuse and block launch order}
Recall that thread blocks need to communicate via HBM. As introduced in \Cref{sec2_preliminaries}, when thread blocks reuse memory, the data is often available in L2 cache, which is significantly faster than HBM. However, cache eviction means that these reuse qualities depend on the order in which blocks get launched. 
For our attention and GEMM kernels, we measure efficiency as we vary block order, summarized in \Cref{tab:l2_table}. Block order substantially affects L2 reuse (measured through HBM bandwidth), which in turn can control kernel performance.

\begin{table}[]
\centering
\begin{minipage}{.48\textwidth}
  \centering
  \begin{tabular}{ccc}
    \toprule
    \multicolumn{3}{c}{\textbf{Matrix Multiply (M=N=K=16384)}} \\
    \midrule
    Block Order & HBM GB/s & TFLOPS \\ 
    \midrule
    \{8, N, M/8\} & 982 & 805 \\ 
    \{N, M\} & 3,070 & 392 \\ 
    \bottomrule
  \end{tabular}
\end{minipage}
\hfill
\begin{minipage}{.48\textwidth}
  \centering
  \begin{tabular}{ccc}
    \toprule
    \multicolumn{3}{c}{\textbf{Attention Forward (D=128)}} \\
    \midrule
    Block Order & HBM GB/s & TFLOPS \\ 
    \midrule
    \{N, H, B\} & 213 & 600 \\ 
    \{B, H, N\} & 2,390 & 494 \\ 
    \bottomrule
  \end{tabular}
\end{minipage}%
\caption{\textbf{L2 reuse} We vary the block orders and measure both consumed bandwidth from HBM (GB/s) and efficiency (TFLOPs). For attention, we consider an optimized kernel, with an internal tiling of 8 rows of blocks, versus a naive kernel that schedules blocks in row-major order. For attention, we compare block order (1) sequence length $N$, heads $H$, and outermost batch $B$ vs. (2) innermost $B$, $H$, then outermost $N$. Different block orders have significant performance implications.}
\vspace{-3mm}
\label{tab:l2_table}
\end{table}

\vspace{-2mm}
\section{Experiments}
\vspace{-1mm}
\label{sec:sec4_results}

In experiments, we validate that \name{} speeds up a broad range of ML primitives. We compare to well-optimized kernels from prior work, written in alternate frameworks such as CutLass, CuBLAS, general CUDA, and Triton.
We compare our kernels for the ``workhorse'' operations in AI, GEMM and attention, as well as kernels for emerging AI architectures, such as linear attention and state space models (\Cref{sec_4:workhorse_kernels}). We profile the kernels to understand \shortname{}'s role in achieving high performance in \Cref{sec4:exotic_kernels}. Kernel listings, in the \shortname{} template, are in \Cref{app:kernel_listings}.
\clearpage

\begin{wrapfigure}{rt}{0.5\linewidth}
\centering
\vspace{-4mm}
\includegraphics[width=0.95\linewidth]{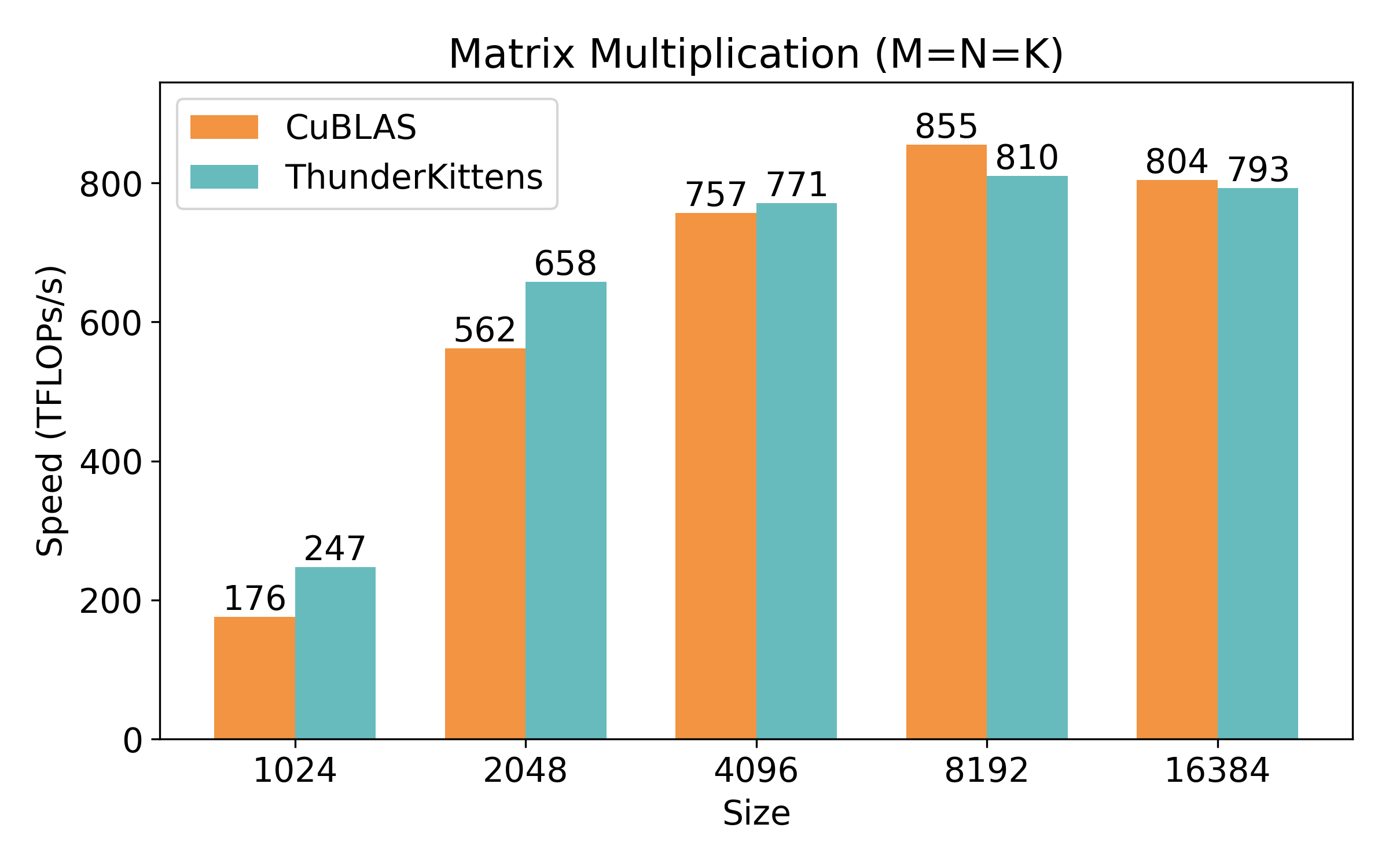}
\vspace{-4mm}
\caption{GEMM kernel from CuBLAS and \shortname{}.}
\vspace{-8mm}
\label{fig:matmul}
\end{wrapfigure}

\vspace{-3mm}
\subsection{\shortname{} enables simple and performant AI kernels}
This section shows a suite of kernels that we develop in the \shortname{} framework.  We benchmark the kernels on NVIDIA H100 80GB SXM GPUs using CUDA 12.6 and report the average TFLOPS. 

 \paragraph{Workhorse kernels for AI}
\label{sec_4:workhorse_kernels}
Industry teams and researchers have made significant investments into optimizing GEMMs and attention over the past several years \citep[inter alia.]{nvidia2023cublas, dao2022flashattention, bikshandi2023acase, dao2024flashattention3}, two workhorse operations that power the Transformer architecture \citep{vaswani2018attention}.
Despite the investment, \shortname{} kernels written entirely in the \shortname{} abstractions and LCSF template can match or outperform the strongest baselines:

\begin{itemize}[itemsep=0.1pt,topsep=0pt,leftmargin=*]
    \item \textbf{GEMM} We compare to the strongest available baseline: CuBLAS for GEMMs \citep{nvidia2023cublas}. We show a single matrix multiply kernel, with just 40 lines of device code, can compete with CuBLAS

    \item \textbf{Attention.} We support multiple variants of attention: causal, non-causal, and grouped query attention~\citep{ainslie2023gqa} at head dimensions $64$ and $128$. We compare to the strongest available baseline, which is concurrent to our work: FlashAttention-3 (FA3)~\citep{dao2024flashattention3}. \shortname{} competes with FA3 across sequence lengths on the non-causal forwards pass, and outperforms FA3 on the causal and non-causal backwards pass by over $40\%$ at short sequences and $10\%$ at longer sequences. 
\end{itemize}

\begin{figure}
\centering
\includegraphics[width=0.47\linewidth]{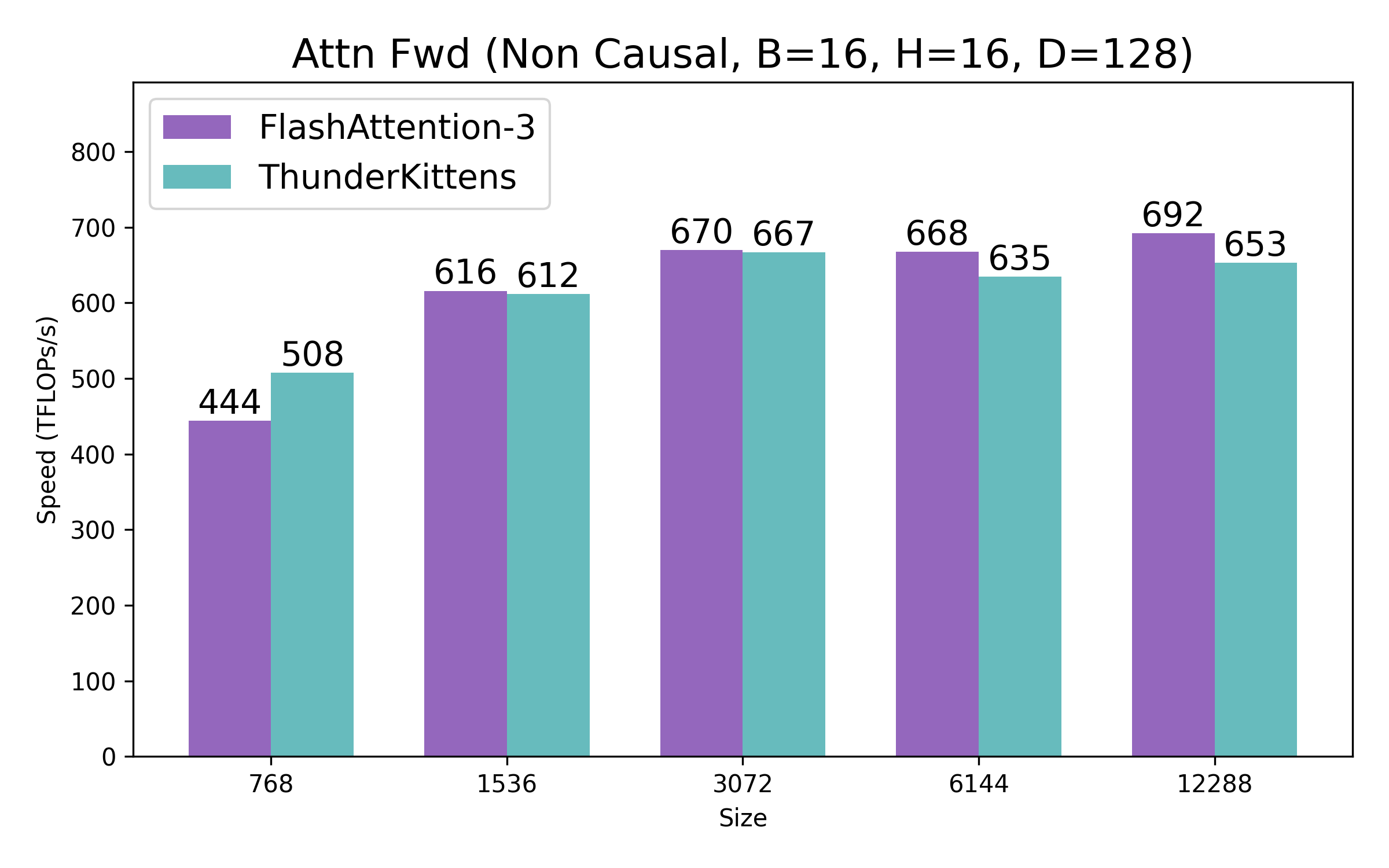}
\hfill
\hspace{-3mm}
\includegraphics[width=0.47\linewidth]{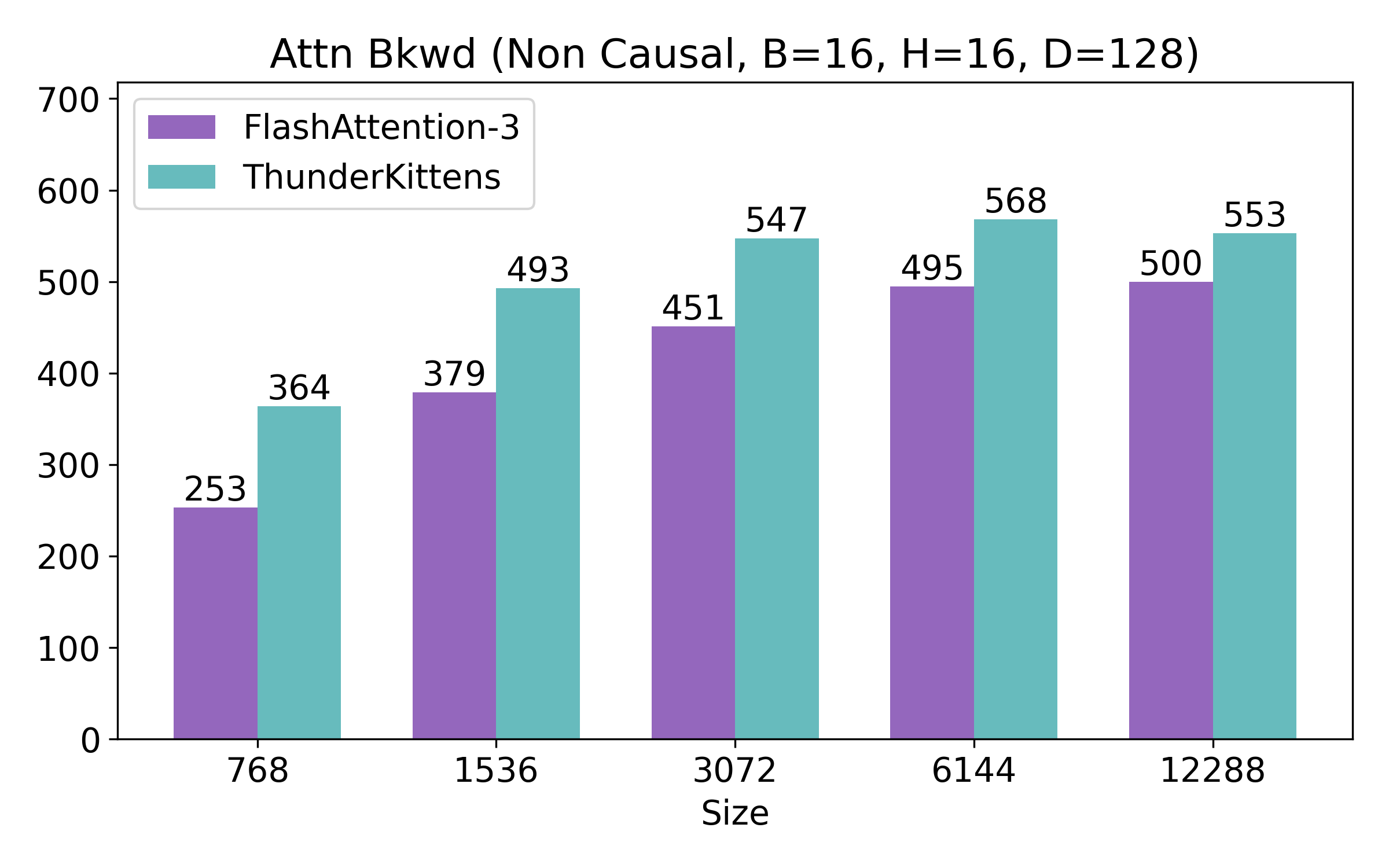}
\vspace{-4mm}
\includegraphics[width=0.47\linewidth]{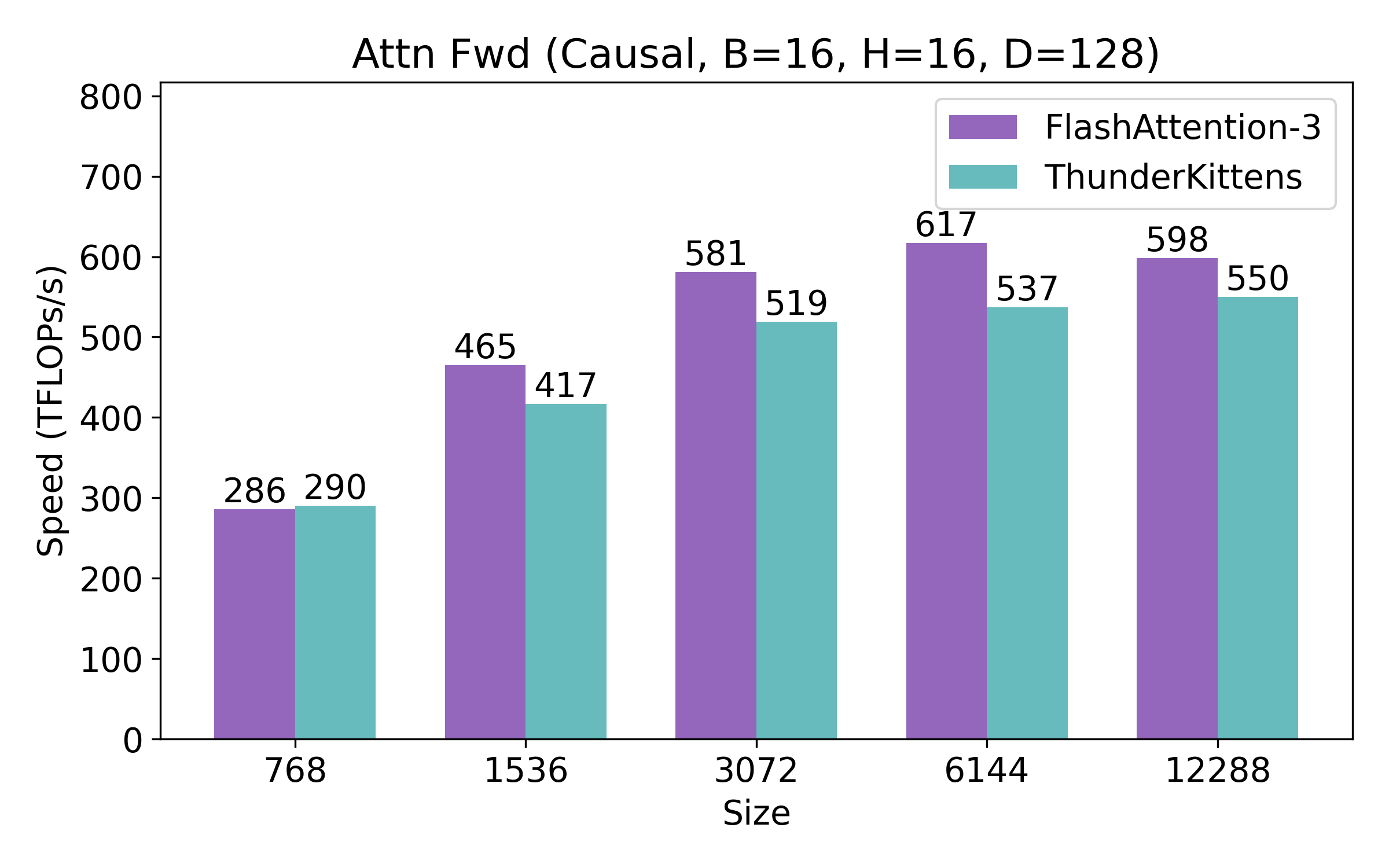}
\hfill
\hspace{-3mm}
\includegraphics[width=0.47\linewidth]{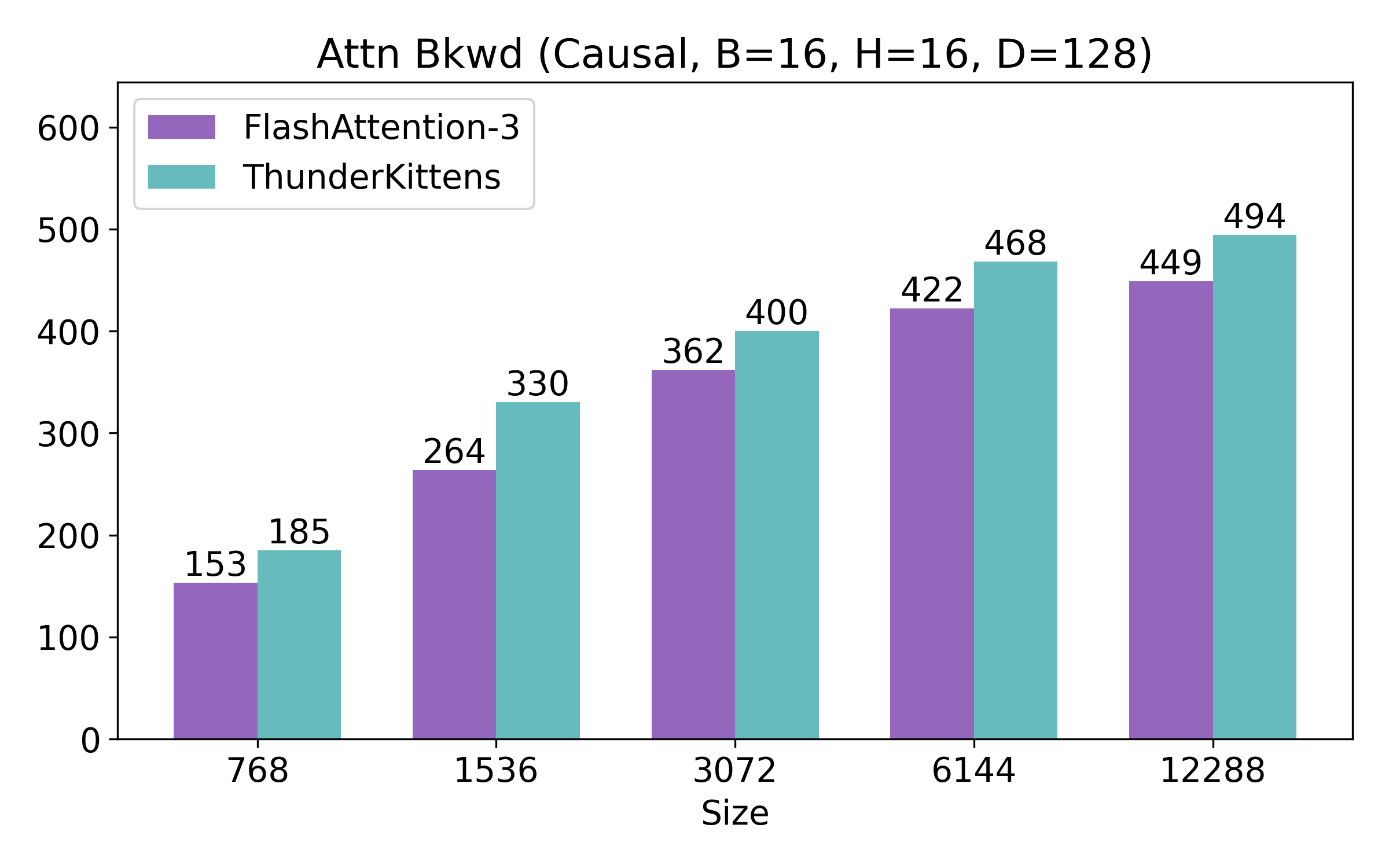}
\vspace{-2mm}
\caption{Attention causal and non causal inference and backwards pass efficiency.
}
\label{fig:matmul_and_attention}
\end{figure}

We find that \shortname{} makes it easy to use the GPU effectively by simplifying the choice of memory layouts, exploration of grid patterns for L2 reuse, and selection of occupancy and pipeline depth. The baseline kernels successfully use specialized H100 instructions and manage memory. However, the existing kernels are relatively complex: FlashAttention-3 proposes a ``ping-pong scheduler'' for workers, and the CuBLAS library is $>$600MB in CUDA 12.6 (\Cref{tab:libraries}), containing many tuned GEMM variants and logic to select the best option at runtime~\citep{schuetze2024reverse}. 
With \shortname{}, we remove the ping-pong and maintain FA3-level efficiency, and we compete with CuBLAS on the demonstrated matrix sizes, using a single GEMM kernel (entirely in \Cref{app:listing_gemm}).

\vspace{-3mm}
\paragraph{Kernels for emerging AI architectures} In addition to supporting peak performance on popular operations like GEMMs and attention, \shortname{} is also designed for to be extensible to emerging AI workloads.
We release a family of kernels across newer machine learning primitives, including linear attention \citep{katharopoulos-et-al-2020}, FFT convolutions \citep{cooley1965algorithm}, and state space models \citep{gu2021efficiently}. 

\begin{figure}[]
\centering
\vspace{-2mm}
\begin{minipage}[b]{0.48\linewidth}
    \centering
    \includegraphics[width=\linewidth]{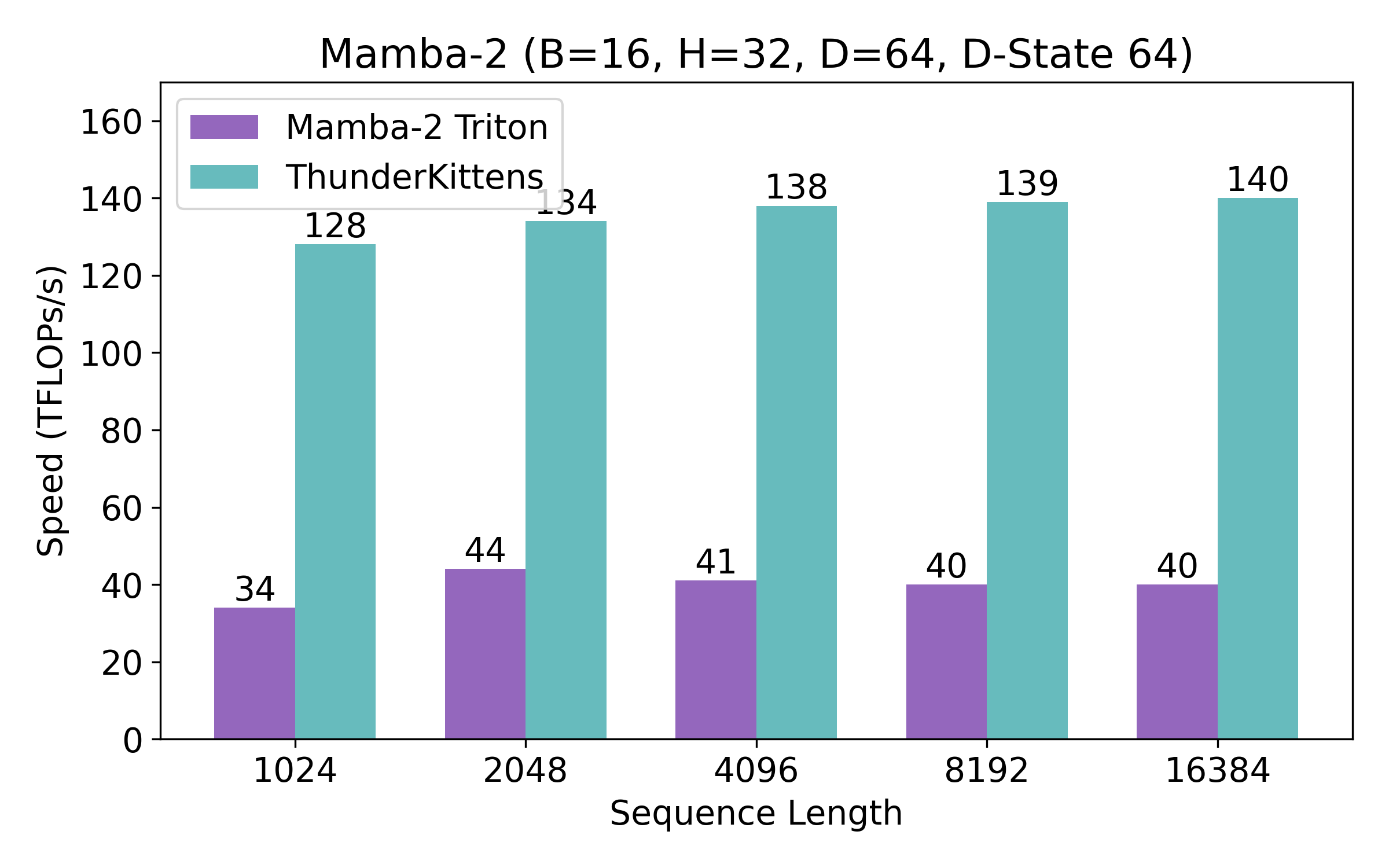}
\end{minipage}
\hfill
\begin{minipage}[b]{0.48\linewidth}
    \centering
    \includegraphics[width=\linewidth]{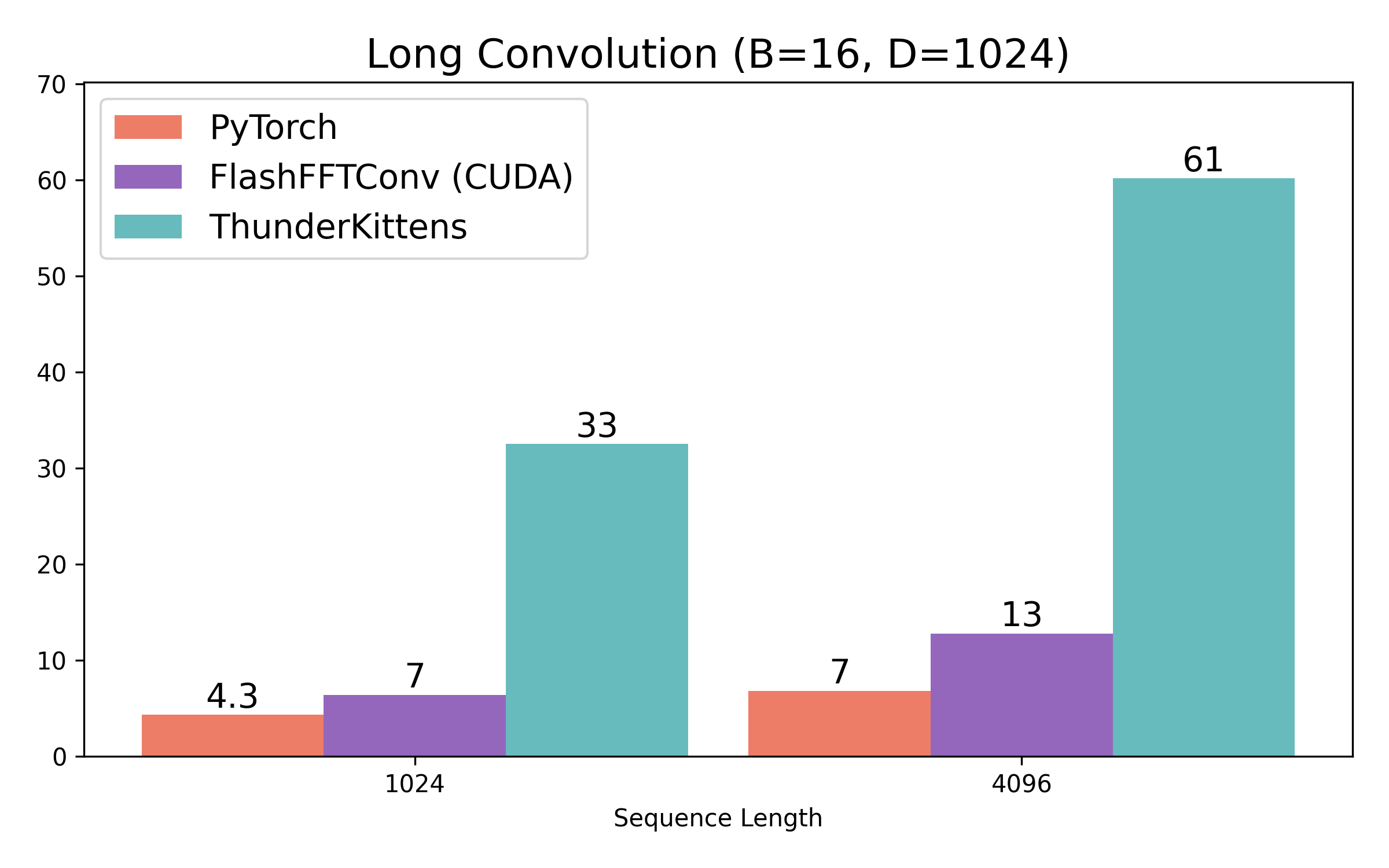}
\end{minipage}
\hfill
\hspace{-3mm}
\begin{minipage}[b]{0.48\linewidth}
    \centering
    \includegraphics[width=\linewidth]{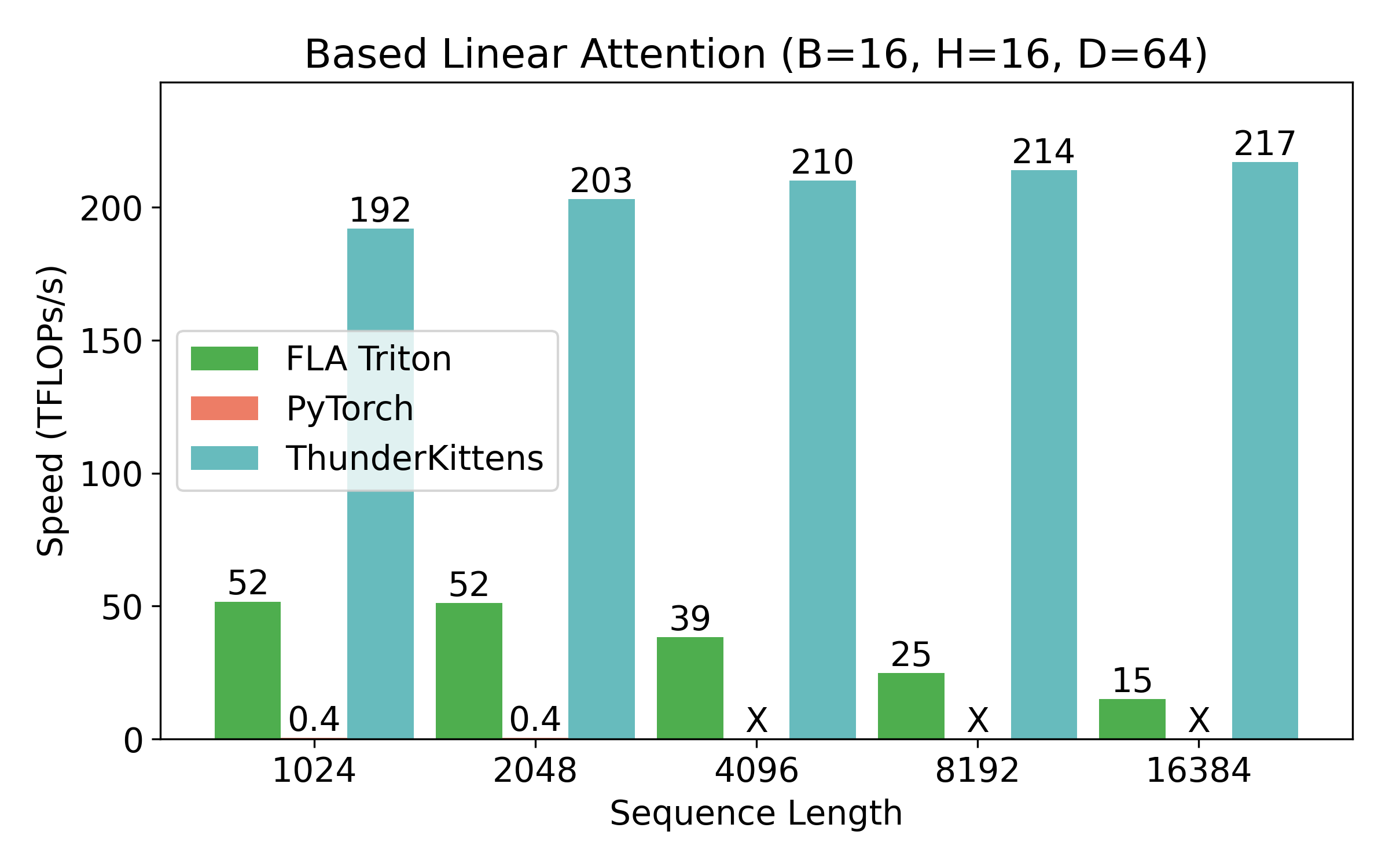}
\end{minipage}
\hfill
\hspace{-2mm}
\begin{minipage}[b]{0.48\linewidth}
    \centering
    \includegraphics[width=\linewidth]{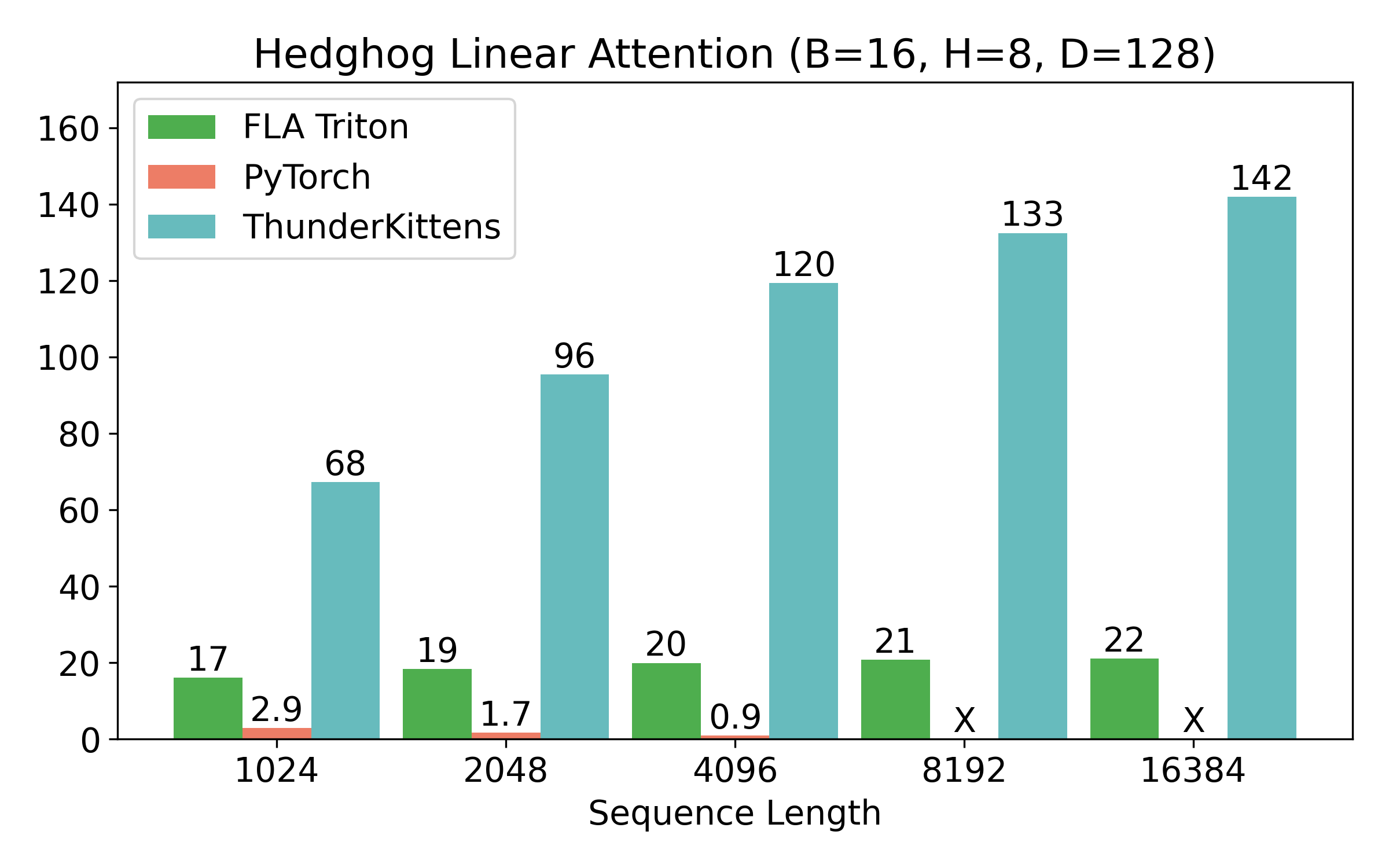}
\end{minipage}
\vspace{-2mm}
\begin{minipage}[b]{0.48\linewidth}
    \centering
    \includegraphics[width=\linewidth]{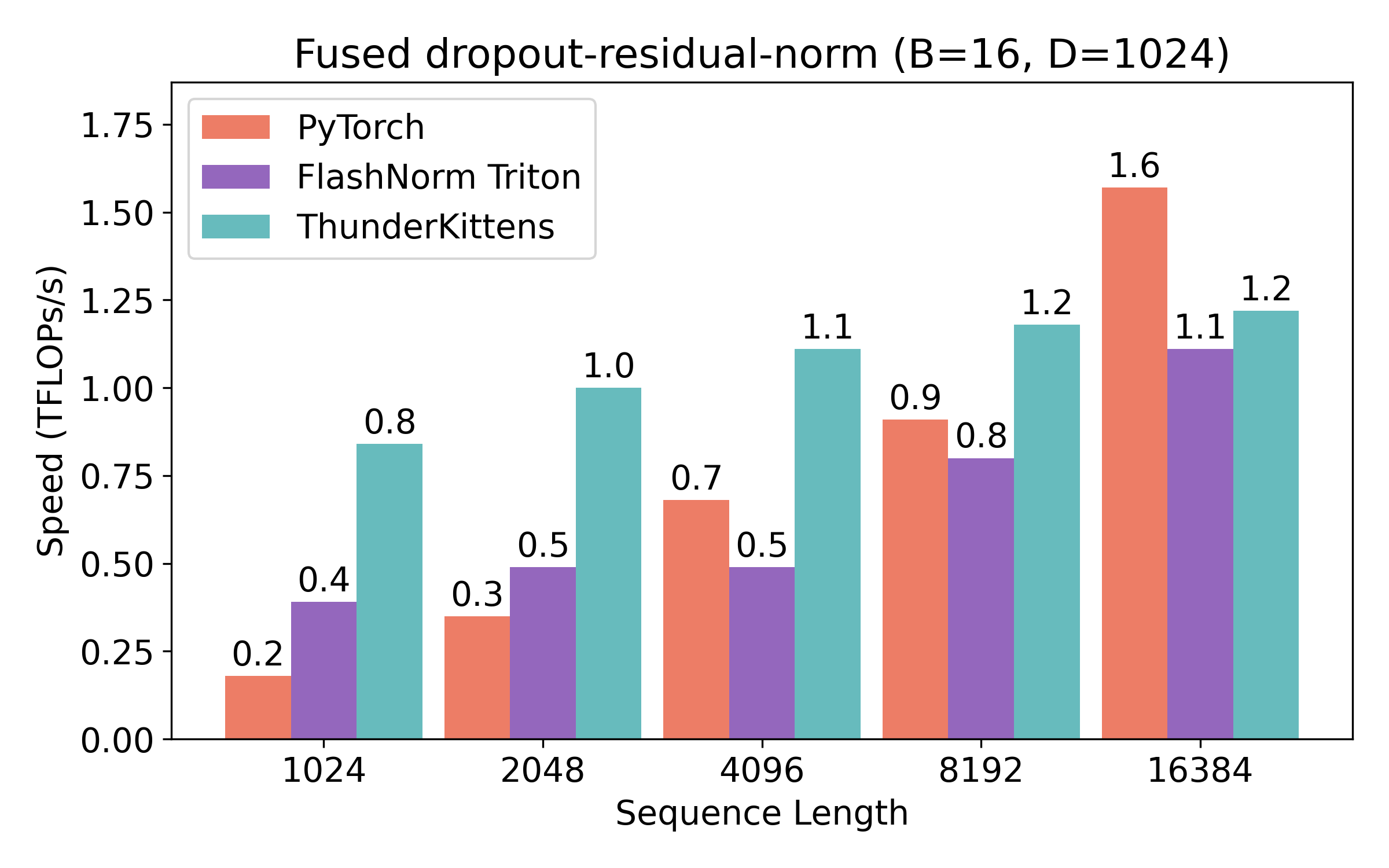}
\end{minipage}
\hfill
\begin{minipage}[b]{0.48\linewidth}
    \centering
    \includegraphics[width=\linewidth]{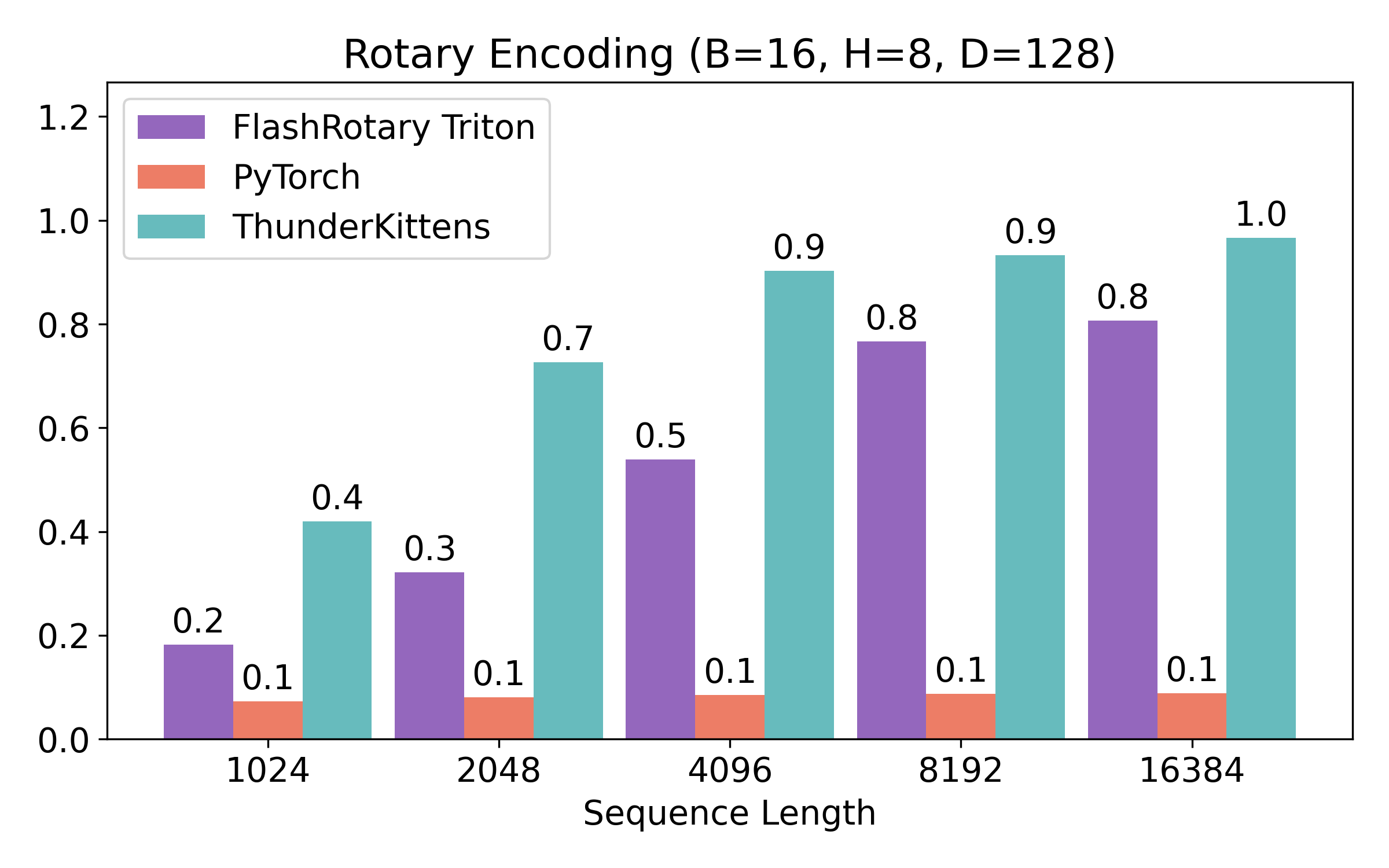}
\end{minipage}
\vspace{-4mm}
\caption{ThunderKittens kernels are performant across a wide range of kernels. We benchmark the kernels on an 80GB NVIDIA H100 GPU and report TFLOPs.}
\label{fig:exotic_kernels}
\end{figure}

\begin{itemize}[itemsep=0.1pt,topsep=0pt,leftmargin=*]
    \item \textbf{Linear attention} We optimize two different classes of linear attention architectures, polynomial-based feature maps as in \citep{arora2024simple, rebased, keles2023on, kacham2024polysketchformerfasttransformerssketching} and \textit{learned} feature maps as in \citep{zhang2024hedgehog, zhang2024lolcats}. In \Cref{fig:exotic_kernels}, we compare to the strongest available baselines: the popular Flash Linear Attention (FLA) CUDA kernels~\citep{yang2024fla}, which are written in Triton. We show \shortname{} outperforms FLA's polynomial-based linear attention by $14\times$. \shortname{} outperforms FLA's learned map linear attention by $6.5\times$.
    \item \textbf{State space models} The long convolution, implemented with Fourier transforms using the convolution theorem,  is the key primitive in popular state space modeling architectures such as S4, H3, and Hyena \citep[inter alia.]{gu2021efficiently, poli2023hyena, dao2022hungry, hasani2022liquid, agarwal2024spectralstatespacemodels}. In \Cref{fig:exotic_kernels}, we compare to the strongest available baseline: the FlashFFTConv CUDA kernels in \citet{fu2023flashfftconv} and show \shortname{} outperforms the prior work by $4.7\times$ at sequence length $4096$ and $7.9\times$ at $1024$. \shortname{} outperforms PyTorch's FFT operations by up to $8.7\times$.
     
     We also optimize the recent Mamba-2 state space model~\citep{dao2024transformers}. We provide a \shortname{} kernel that outperforms the Triton kernels in prior work ~\citet{dao2024transformers} by $>3\times$  (\Cref{fig:exotic_kernels}). This gap is primarily due to the ease of fusing complex operations in \shortname{}.
\end{itemize}

The abstractions in \shortname{} -- including specialized instructions such as TMA and WGMMA, and register tiles to manage register memory effectively -- contribute to these large improvements. The baseline kernels do not use these GPU features.

\vspace{-3mm}
\paragraph{\shortname{}'s programming model is extensible.}
We develop kernels for common memory AI operations -- fused dropout-residual-layernorm~\citep{ba2016layer}, and rotary~\citep{su2023roformer} -- and show \shortname{} is effective. We compare to popular Triton kernels for these operations. \footnote{Reference trton kernels are from \url{https://github.com/Dao-AILab/flash-attention}.}

\subsection{Comparing kernel implementations}
\label{sec4:exotic_kernels}

\begin{table}[]
\centering
\vspace{-2mm}
\begin{tabular}{lccccc}
\toprule
& \multicolumn{2}{c}{Occupancy utilizations (\%)} & \multicolumn{2}{c}{HBM} & \multicolumn{1}{c}{Shared} \\ 
\cmidrule(r){2-3} \cmidrule(r){4-5} \cmidrule(r){6-6}
Impl. & Tensor core & Issue slots & TPS (GB/s) & Stalls (Cycles) & Stalls (Cycles) \\ 
\midrule
FA3  Bkwd        & 61.2 & 25.1 & 328 & 1.83 & 0.92  \\ 
\shortname{} Bkwd & 58.2 & 34.8 & 490 & 1.63 & 0.14  \\ 
\midrule
FlashFFT &  13.4 & 25.5 & 14.8 & 2.5 & 1.6 \\ 
\shortname{} &  54.8 & 40.0 & 31.4 & 0.6 & 0.3  \\ 
\bottomrule
\end{tabular}
\vspace{-3mm}
\caption{Profiles for 1) attention backwards pass kernels from FlashAttention-3~\citep{dao2024flashattention3} vs. \shortname{} and 2) long convolution kernels from FlashFFTConv~\citep{fu2023flashfftconv} vs. \shortname{}, obtained using NVIDIA NSight Compute. }
\label{tab:ncu_profiles}
\end{table}

To further compare \shortname{} and baseline kernels,  we profile the kernels using NVIDIA's NSight Compute (NCU) tool.
\footnote{\url{https://developer.nvidia.com/nsight-compute}} In \Cref{tab:ncu_profiles}, we give NCU profiles for both the emerging long convolution primitive and the well-optimized attention backwards pass, comparing to the strongest respective baselines.
\vspace{1mm}
\begin{itemize}[itemsep=0.1pt,topsep=0pt,leftmargin=*]
    \item \textbf{Long convolution} We profile FlashFFTConv (FC) and \shortname{} long convolution kernels at $B,D,N=16,1024,4096$ in NCU. We find \shortname{} helps both with overlapping the workers (indicated by higher issue slots and fewer memory stalls) and in tensor core utilization ($4.1\times$ increase). This is enabled by our \shortname{} template, and use of \shortname{} warpgroup operations (which saves registers and establishes a SMEM to register memory pipeline through warpgroup matrix-multiply-add (WGMMA) operations).
    \item \textbf{Attention backwards} We consider FA3 and \shortname{} at $B,H,N,D=16,16,3072,128$. The methods match in tensor core utilization, but \shortname{} gives higher issue slot utilization, suggesting the occupancy may be better-tuned. In HBM costs, \shortname{} gives higher memory throughput and correspondingly incurs $10\%$ fewer stalled cycles on HBM waits. For shared memory, \shortname{} incurs $85\%$ fewer stalled cycles -- we find \shortname{} has \textit{no bank conflicts}, but NVIDA's NCU profiler reports up to $9.6$-way bank conflicts in FA-3.
\end{itemize}
\vspace{1mm}
The kernel profiles highlight the difficulty of simultaneously managing each type of GPU parallelism, and we hope \shortname{} can help reduce the effort. 
We provide example \shortname{} kernels listings in \Cref{app:kernel_listings}.

\section{Conclusion}
Given the challenge of mapping AI architectures to GPU hardware, our work asks how far we can get with a few easy to use GPU programming abstractions. In \name{}, we give an abstraction for each level of the GPU hierarchy: tiles with managed layouts at the worker level and a asynchronous execution LCSF template at the thread block level. We highlight options and tradeoffs for persistent block launches and L2 reuse at the grid level. The natural question is whether we sacrifice anything in performance when we write kernels with so few abstractions. We implement a breadth of AI kernels in \shortname{} and excitingly find that our abstractions are both general and consistently meet or exceed state-of-the-art. We are optimistic about the potential for simple and accessible ways of programming AI hardware. 

To support future development, our framework and kernels are open sourced at: \url{https://github.com/HazyResearch/ThunderKittens}.

\section{Acknowledgements}
We are grateful to Together.ai for making this work possible. 
We than Arjun Parthasarthy for assistance in developing the complex tile support and FlashFFT kernels. 
We thank Mayee Chen, Tri Dao, Kawin Ethyarajh, Sabri Eyuboglu, Neel Guha, David Hou, Jordan Juravsky, Hermann Kumbong, Jerry Liu, Avner May, Quinn McIntyre, Jon Saad-Falcon, Vijay Thakkar, Albert Tseng, Michael Zhang for helpful feedback and discussions during this work. 
We gratefully acknowledge the support of NIH under No. U54EB020405 (Mobilize), NSF under Nos. CCF2247015 (Hardware-Aware), CCF1763315 (Beyond Sparsity), CCF1563078 (Volume to Velocity), and 1937301 (RTML); US DEVCOM ARL under Nos. W911NF-23-2-0184 (Long-context) and W911NF-21-2-0251 (Interactive Human-AI Teaming); ONR under Nos. N000142312633 (Deep Signal Processing); Stanford HAI under No. 247183; NXP, Xilinx, LETI-CEA, Intel, IBM, Microsoft, NEC, Toshiba, TSMC, ARM, Hitachi, BASF, Accenture, Ericsson, Qualcomm, Analog Devices, Google Cloud, Salesforce, Total, the HAI-GCP Cloud Credits for Research program, the Stanford Data Science Initiative (SDSI), BFS is supported by a Hertz Fellowship, SA is supported by a SGF Fellowship, and members of the Stanford DAWN project: Meta, Google, and VMWare. The U.S. Government is authorized to reproduce and distribute reprints for Governmental purposes notwithstanding any copyright notation thereon. Any opinions, findings, and conclusions or recommendations expressed in this material are those of the authors and do not necessarily reflect the views, policies, or endorsements, either expressed or implied, of NIH, ONR, or the U.S. Government.

\bibliography{main}
\bibliographystyle{main}
\appendix
\clearpage
\noindent Our appendix is organized as follows: 
\vspace{1mm} 
\begin{enumerate}[itemsep=0.1pt,topsep=0pt,leftmargin=*]
    \item \Cref{app:related_work} provides an extended discussion of related work.
    \item \Cref{app:kernel_listings} provides a set of kernel listings written in the \shortname{} abstractions. 
    \item \Cref{app:layouts} provides extended discussion of approaches to constructing shared memory layouts and the set of layouts used in \shortname{}.
\end{enumerate}

\section{Related work}
\label{app:related_work}
We provide an initial discussion of related work in \Cref{sec2_preliminaries} and extended discussion here. We discuss frameworks that support AI kernel development and prior work to develop hardware-aware AI algorithms.

\paragraph{CPP embedded libraries} Towards raising the level of abstraction and supporting simpler programs, NVIDIA maintains the CuTe and CUTLASS libraries, which are \textit{CUDA} primitives library for graphics, scientific computing, and ML. As discussed in \Cref{sec2_preliminaries}, both CUTLASS and \shortname{} can support the same kernels, since both are C++ embedded libraries. Developers can use the power of the full C++ library, including raw CUDA, when using these frameworks. Distinct from CUTLASS's objectives, we specifically explore how broad and fast we can go, just using a small number of opinionated abstractions.

\paragraph{Compiler-based libraries} Many machine learning frameworks employ high-level computational graph representations for optimizations, such as TVM \cite{tvm}, TensorFlow XLA \cite{tensorflow}, Glow \cite{glow}, and DLVM \cite{dlvm}. TVM, for instance, incorporates a flexible tensor expression language and automated schedule optimization. Building upon Halide's \cite{halide} separation of algorithm and schedule, TVM introduces new primitives such as tensorization. These frameworks' approach differs from \name{} in that they provide a full end-to-end stack that optimizes both graph-level and operator-level transformations, while \name{} concentrates specifically on kernel-level optimizations.

Triton \cite{triton} builds on existing approaches to deep learning compilation while introducing novel techniques for efficient tiled computations on GPUs. Recently, tools such as Flex Attention also provide easy to use interfaces to write kernels for attention variants, and compile down to Triton~\cite{he2024flex}.
Unlike XLA and Glow, which use tensor-level IRs and predefined templates, or Tensor Comprehensions \cite{tensorcomprehensions} and Diesel \cite{diesel}, which rely on polyhedral models, Triton introduces a C-like language (Triton-C) and an LLVM-based IR (Triton-IR) centered around parametric tile variables. This approach supports non-affine tensor indices that polyhedral models struggle with. Triton's JIT compiler (Triton-JIT) implements tile-level optimizations and an auto-tuner, often enabling high performance on par with hand-tuned libraries. A key difference between \shortname{} and Triton is that \shortname{} is embedded within CUDA, and as a result its abstractions fail gracefully. In contrast, operations in Triton must be written entirely within its framework.

\paragraph{Hardware-aware AI architectures} We are inspired by the success of several prior works that introduce systems-level innovations to improve ML efficiency such as FlashAttention~\citep{dao2022flashattention, dao2023flashattention2, dao2024flashattention3} and other optimized attentions~\citep{bikshandi2023acase}, FlashFFTConv~\citep{fu2023flashfftconv}, linear attention kernels~\cite{vyas_et_al_2020, yang2024fla, peng2023rwkv, dao2024transformers}. 
Given the effort required to independently optimize each new architecture, we ask whether a small set of opinionated GPU programming abstractions can obtain kernels for a broad range of AI operations.

\begin{table}[b]
\centering
\vspace{-2mm}
\begin{tabular}{lcc}
\toprule
Library & Size (Bytes) & Date / Version  \\ 
\midrule
CutLASS  & 22MB  & 10/22/2024 \\ 
CuBLAS   & 689MB  & CUDA 12.6 \\ 
Triton   & 12.6MB  & 10/22/2024 \\ \hline
\shortname{} & $<$1.0MB  & 10/22/2024 \\ 
\bottomrule
\label{tab:libraries}
\end{tabular}
\vspace{-3mm}
\caption{Sizes of various CUDA libraries. For CuTLASS and \shortname{} we report the size of the ``include/'' directory, for CuBLAS we inspect the sizes of the static library files in CUDA, and for Triton we report the combined size of the ``include/'' directories in Triton and in the core MLIR compiler dependency.}
\end{table}
\clearpage 
\section{\name{} kernels}
\label{app:kernel_listings}

This section first recaps our benchmarking methodology for the results and provides a set of kernels written in the \shortname{} LCSF template and tile abstractions:
\begin{enumerate}
    \item \Cref{app:listing_gemm} GEMM kernel
    \item \Cref{fig:fft-pc-template} Long convolution kernel
    \item \Cref{fig:attn-pc-template} Attention kernel
    \item \Cref{fig:rotary-pc-template} Rotary kernel
\end{enumerate}
To introduce the template components, we describe the GEMM kernel in detail in \Cref{app:listing_gemm}. 

\paragraph{Benchmarking approach} Our kernels in \Cref{sec:sec4_results} are benchmarked on an NVIDIA H100 80GB SXM GPU with 10 warmup and 10 timed iterations using timings measured in C++. We also provide Python-bound kernels and benchmarking infrastructure in our repository for reference. 

\subsection{Matrix multiply}
\label{app:listing_gemm}
First we show and describe a \shortname{} GEMM kernel in the LCSF template. 

Each compute warpgroup is responsible for computing $64M$-row, $64N$-column chunk of the resulting output matrix. Each compute worker identifies the coordinates for its chunk, zeros its accumulator registers, repeatedly runs large asynchronous matrix multiplies (compute), and finally stores out its tile in the end (finish). The load workers also compute their coordinates, and then repeatedly load chunks of the input matrices (load). Store workers perform asynchronous stores when the compute workers are finished with the chunks (stores).

\paragraph{Tuning the number of workers and pipeline stages} The computation is divided into stages, with each stage processing 64 elements along the reduction dimensions of the input matrices. The input pipeline is automatically sized by \name{} if the user does not specify a value. For common configurations of either a (2 compute warpgroup) $128\times256$ or ($3$ compute warpgroups) $192\times192$ output tile per block, it generates a 4-stage pipeline.

\paragraph{Tuning the grid order} The greatest complexity of this kernel is in setting the grid parameters. This kernel adopts a 3D stride over the input matrices, which has a significant effect for large matrices which do not fit in L2 cache. The order in which blocks execute strongly influences cache locality and thus available memory bandwidth. To illustrate the magnitude of the effect, comparing the presented scheme versus a naive grid (in which blocks are executed in row-major order) a $4096\times4096\times4096$ matrix multiply only drops from $767$ TFLOPs to $735$ TFLOPs, but a $16384\times16384\times16384$ matrix multiply drops from $797$ TFLOPs to $387$ TFLOPs, a $>50\%$ performance degradation.

\begin{figure}[b]
\begin{lstlisting}[basicstyle=\scriptsize\ttfamily]
using namespace kittens;
using namespace kittens::prototype;
using namespace kittens::prototype::lcf;
template<int M_BLOCK, int N_BLOCK>
struct matmul_layout {
  using  base_tile      = st_bf<64, 64>;
  using  global_layout  = gl<bf16, 1, 1, -1, -1, base_tile>;
  struct globals        { global_layout A, B, C; };
  struct input_block    { base_tile a[M_BLOCK], b[N_BLOCK]; };
  struct finish_block   { base_tile c[M_BLOCK][N_BLOCK]; };
  struct common_state   { int2 coord; };
  struct consumer_state { rt_fl<16, N_BLOCK*base_tile::cols> accum; };
};
\end{lstlisting}
\end{figure}

\begin{figure}
\begin{lstlisting}[basicstyle=\scriptsize\ttfamily]
template<int _M_BLOCK=2, int _N_BLOCK=4, int _SUPER_M=12>
struct matmul_template {
  static constexpr int M_BLOCK = _M_BLOCK, N_BLOCK = _N_BLOCK, SUPER_M = _SUPER_M;
  using layout    = matmul_layout<M_BLOCK, N_BLOCK>;
  using wide_tile = st_bf<64, 64*N_BLOCK>;
  static constexpr int NUM_CONSUMER_WARPS=M_BLOCK*4, INPUT_PIPE_STAGES=4, PRODUCER_BARRIER_ARRIVALS=1;
  // Helper functions
  template<bool PERISISTENT_GRID=true> __host__ static inline dim3 grid(int M, int N, int K) {
    return dim3(PERISISTENT_GRID ? 132 : M*N/(M_BLOCK*N_BLOCK*layout::base_tile::num_elements));
  }
    // ThunderKittens template functions
  __device__ static inline void common_setup(common_setup_args<layout> args) {
    int Rblocks = args.globals.C.rows / (M_BLOCK*64), Cblocks = args.globals.C.cols / (N_BLOCK*64);
    int super_rows = (Rblocks/SUPER_M)*SUPER_M,
      final_rows = Rblocks - super_rows,
      super_repeat = SUPER_M*Cblocks;
    int task_id = args.task_iter*gridDim.x + blockIdx.x;
    if (task_id < super_rows * Cblocks)
      args.common.coord = { SUPER_M*(task_id/super_repeat) + task_id%SUPER_M,
                            (task_id%super_repeat)/SUPER_M };
    else if (task_id < Rblocks*Cblocks) {
      int remainder_id = task_id - super_rows*Cblocks;
      args.common.coord = { super_rows + (remainder_id%final_rows), remainder_id/final_rows };
    }
    else { // Id is too high, no more work to do
      args.num_iters = -1;
      return;
    }
    args.num_iters = args.globals.A.cols/64;
    int id = warpgroup::groupid() == NUM_CONSUMER_WARPS/4 ? 0 : warpgroup::groupid(); // producer sets as 0
    args.common.coord = { args.common.coord.x*M_BLOCK + id, args.common.coord.y*N_BLOCK };
  }
  struct producer {
    __device__ static void setup(producer_setup_args<layout> args) {
      warpgroup::decrease_registers<40>(); // decrease registers for producers
    }
    __device__ static void load(producer_load_args<layout> args) {
      if(warpgroup::warpid() == 0) {
        tma::expect(args.inputs_arrived, args.input);
        for(int i = 0; i < M_BLOCK; i++)
          tma::load_async(args.input.a[i], args.globals.A,
                          {args.common.coord.x+i, args.iter}, args.inputs_arrived);
        for(int i = 0; i < N_BLOCK; i++)
          tma::load_async(args.input.b[i], args.globals.B,
                          {args.iter, args.common.coord.y+i}, args.inputs_arrived);
      }
    }
  };
  struct consumer {
    __device__ static void setup(consumer_setup_args<layout> args) {
      warpgroup::increase_registers<232>(); // increase registers for consumers
      zero(args.state.accum);
    }
    __device__ static void compute(consumer_compute_args<layout> args) {
      warpgroup::mma_AB(
        args.state.accum, // dest registers
        args.input.a[warpgroup::groupid()], // A matrix
        reinterpret_cast<wide_tile&>(args.input.b) // B matrix
      );
      warpgroup::mma_async_wait();
      if(laneid() == 0) arrive(args.inputs_finished);
    }
    __device__ static void finish(consumer_finish_args<layout> args) {
      warpgroup::store(reinterpret_cast<wide_tile&>(args.finish.c[warpgroup::groupid()]), args.state.accum);
      warpgroup::sync();
      if(warpgroup::warpid() == 0) for(int i = 0; i < N_BLOCK; i++) {
        tma::store_async(args.globals.C, args.finish.c[warpgroup::groupid()][i],
                                   {args.common.coord.x, args.common.coord.y+i});
        tma::store_async_read_wait(); // wait that store is finished before reusing finish memory
      }
      zero(args.state.accum);
      if(laneid() == 0) arrive(args.finish_finished);
    }
  };
};
\end{lstlisting}
\label{fig:matmul-pc-template}
\caption{Templated matrix multiply kernel which is reasonably competitive with CuBLAS.}
\end{figure}

\clearpage
\subsection{Long convolution}
This section shows the long convolution kernel for sequence length $4096$, written in the \shortname{} abstractions.
We use the FFT convolution algorithm, computed via Monarch Matrices, for our long convolution kernel~\citep{cooley1965algorithm, fu2023monarch, dao2022monarchexpressivestructuredmatrices}. 

\begin{figure}[h!]
\begin{lstlisting}[basicstyle=\scriptsize\ttfamily]
using namespace kittens;
using namespace kittens::prototype;
using namespace kittens::prototype::lcsf;
template<int _wg> struct fftconv_4096_layout { // 4096
  static constexpr int wg = _wg;
  using seq_tile      = st_bf<64, 64>;
  using seq_layout    =     gl<bf16, -1, -1, 64, 64, seq_tile>;
  using filter_layout = cgl<gl<bf16,  1, -1, 64, 64, seq_tile>>;
  using fft_layout    = cgl<gl<bf16,  1,  1, 64, 64>>;
  struct globals {
    seq_layout o, x;
    filter_layout kf;
    fft_layout f, finv, tw, twinv_t;
  };
  struct input_block    { seq_tile x[wg]; };
  struct output_block   { seq_tile o[wg]; };
  struct scratch_block  {
    cst_bf<64, 64> kf, f, finv, tw, twinv_t, tmp[2];
  };
  struct consumer_state { int current_head; };
};
struct fft_4096_template {
  static constexpr int NUM_CONSUMER_WARPS=8, NUM_CONSUMER_WARPGROUPS=NUM_CONSUMER_WARPS/4, NUM_BLOCKS=1, OUTPUT_PIPE_STAGES=2, INPUT_PIPE_STAGES=4;
  using layout = fftconv_4096_layout<NUM_CONSUMER_WARPGROUPS>;
  // mine
  __device__ static inline void load_head_data(typename layout::scratch_block &scratch, const layout::globals &g, int head) {
    using consumers = group<NUM_CONSUMER_WARPS>;
    consumers::sync(3);
    consumers::load(scratch.kf, g.kf, {0, head, 0, 0}); // next chunk
    consumers::sync(3);
  }
  // tk
  __device__ static void common_setup(common_setup_args<layout> args) {
    int heads_handled = (args.globals.x.depth+131-blockIdx.x) / 132; // I am guaranteeing batch is handled by just one block.
    int iters_per_head = (args.globals.x.batch + NUM_CONSUMER_WARPGROUPS-1) / NUM_CONSUMER_WARPGROUPS;
    args.num_iters = args.task_iter == 0 ? heads_handled * iters_per_head : -1;
  }
  struct producer {
    __device__ static void setup(producer_setup_args<layout> args) {
      warpgroup::producer_registers();
    }
    __device__ static void load(producer_load_args<layout> args) {
      int iters_per_head = (args.globals.x.batch + NUM_CONSUMER_WARPGROUPS-1) / NUM_CONSUMER_WARPGROUPS;
      int head  = (args.iter / iters_per_head)*132 + blockIdx.x;
      int batch = (args.iter % iters_per_head) * NUM_CONSUMER_WARPGROUPS;
      if(warpgroup::warpid() == args.iter%4) {
        tma::expect_bytes(args.inputs_arrived, sizeof(args.input.x[0]) * min((int)NUM_CONSUMER_WARPGROUPS, (int)(args.globals.x.batch - batch)));
        for(int b = batch; b < batch+NUM_CONSUMER_WARPGROUPS && b < args.globals.x.batch; b++) {
          tma::load_async(args.input.x[b-batch], args.globals.x, { b, head, 0, 0 }, args.inputs_arrived);
        }
        if(laneid() == 0) arrive(args.inputs_arrived, 3); // extra arrivals needed
        __syncwarp();
      }
    }
    __device__ static void store(producer_store_args<layout> args) {
      int iters_per_head = (args.globals.x.batch + NUM_CONSUMER_WARPGROUPS-1) / NUM_CONSUMER_WARPGROUPS;
      int head  = (args.iter / iters_per_head)*132 + blockIdx.x;
      int batch = (args.iter % iters_per_head) * NUM_CONSUMER_WARPGROUPS;
      if(warpgroup::warpid() == args.iter%4) {
        for(int b = batch; b < batch+NUM_CONSUMER_WARPGROUPS && b < args.globals.x.batch; b++) {
          tma::store_async(args.globals.o, args.output.o[b-batch], { b, head, 0, 0 });
        }
        tma::store_async_read_wait();
        if(laneid() == 0) arrive(args.outputs_finished, 4);
        __syncwarp();
      }
    }
  };
\end{lstlisting}
\end{figure}

\begin{figure}[H]
\begin{lstlisting}[basicstyle=\scriptsize\ttfamily]
  struct consumer {
    __device__ static void setup(consumer_setup_args<layout> args) {
      warpgroup::consumer_registers<NUM_CONSUMER_WARPS/4>();
      int iters_per_head = (args.globals.x.batch + NUM_CONSUMER_WARPGROUPS-1) / NUM_CONSUMER_WARPGROUPS;
      args.state.current_head = (0 / iters_per_head)*132 + blockIdx.x; // start for iter 0
      using consumers = group<NUM_CONSUMER_WARPS>;
      consumers::load(args.scratch.f,       args.globals.f,       {0, 0, 0, 0});
      consumers::load(args.scratch.finv,    args.globals.finv,    {0, 0, 0, 0});
      consumers::load(args.scratch.tw,      args.globals.tw,      {0, 0, 0, 0});
      consumers::load(args.scratch.twinv_t, args.globals.twinv_t, {0, 0, 0, 0});
      load_head_data(args.scratch, args.globals, args.state.current_head);
    }
    __device__ static void compute(consumer_compute_args<layout> args) {

      int warpgroupid = warpgroup::warpid()/kittens::WARPGROUP_WARPS;
      int default_barrer_id = warpgroupid + 4;

      // X = F^T X
      crt_fl<16, 64> mma_reg; // 64 registers
      crt_bf<16, 64> accum, tmp; // 32 registers each
      warpgroup::mm_AB(mma_reg.real, args.scratch.f.real, args.input.x[warpgroup::groupid()]);
      warpgroup::mm_AB(mma_reg.imag, args.scratch.f.imag, args.input.x[warpgroup::groupid()]);
      warpgroup::mma_async_wait();
      copy(accum, mma_reg);

      warpgroup::load(tmp, args.scratch.tw); // for twiddle first
      mul(accum, accum, tmp);

      group<NUM_CONSUMER_WARPS>::sync(2);
      warpgroup::mm_AB(mma_reg, accum, args.scratch.f);
      warpgroup::mma_async_wait();
      copy(accum, mma_reg);

      warpgroup::load(tmp, args.scratch.kf); // for filter second
      mul(accum, accum, tmp);

      warpgroup::mm_AB(mma_reg, accum, args.scratch.finv);
      warpgroup::mma_async_wait();
      copy(accum, mma_reg);

      warpgroup::load(tmp, args.scratch.twinv_t); // twiddle inverse is pre-transposed
      mul(accum, accum, tmp);

      warpgroup::store(args.scratch.tmp[warpgroup::groupid()], accum); // must store for AtB
      warpgroup::sync(default_barrer_id);

      warpgroup::mm_AB(mma_reg, args.scratch.finv, args.scratch.tmp[warpgroup::groupid()]); // TODO: optimize
      warpgroup::mma_async_wait();
      
      warpgroup::store(args.output.o[warpgroup::groupid()], mma_reg.real); // COMMENT ME OUT LATER
      warpgroup::sync(default_barrer_id);

      if(laneid() == 0) {
        arrive(args.inputs_finished);
        arrive(args.outputs_arrived);
      }
      __syncwarp();
      int iters_per_head = (args.globals.x.batch + NUM_CONSUMER_WARPGROUPS-1) / NUM_CONSUMER_WARPGROUPS;
      int next_head = ((args.iter+1) / iters_per_head)*132 + blockIdx.x;
      if(next_head != args.state.current_head) {
        load_head_data(args.scratch, args.globals, next_head);
        args.state.current_head = next_head;
      }
    }
    __device__ static void finish(consumer_finish_args<layout> args) { if(laneid() == 0) arrive(args.finish_finished); }
  };
};
\end{lstlisting}
\label{fig:fft-pc-template}
\caption{A templated convolution kernel for context length 4096, which outperforms FlashFFTConv~\citep{fu2023flashfftconv}.}
\end{figure}

\clearpage
\subsection{Attention}
\label{fig:atth-pc-template}
This section shows non-causal attention at head dimensions $64, 128$, in the \shortname{} abstractions.

\begin{figure}[h!]
\begin{lstlisting}[basicstyle=\scriptsize\ttfamily]
using namespace kittens;
using namespace kittens::prototype;
using namespace kittens::prototype::lcf;
template<int D, int NUM_WORKERS> struct attn_fwd_layout {
  using qo_tile   = st_bf<64, D>;
  using kv_tile   = st_bf<D==64?192:128, D>;
  using qo_global = kittens::gl<bf16, -1, -1, -1, D, qo_tile>;
  using kv_global = kittens::gl<bf16, -1, -1, -1, D, kv_tile>;
  struct globals { qo_global O, Q; kv_global K, V; };
  struct input_block    { kv_tile k, v; };
  struct scratch_block  { qo_tile q[NUM_WORKERS]; };
  struct common_state   { int batch, head, seq; };
  struct consumer_state {
    rt_fl<16, qo_tile::cols> o_reg;
    col_vec<rt_fl<16, kv_tile::rows>> max_vec, norm_vec;
    col_vec<rt_fl<16, kv_tile::rows>> max_vec_last_scaled, max_vec_scaled;
    rt_fl<16, kv_tile::rows> att_block;
    rt_bf<16, kv_tile::rows> att_block_mma;
  };
};
template<int D> struct attn_fwd_template {
  static constexpr int NUM_CONSUMER_WARPS = 12, NUM_WORKERS = NUM_CONSUMER_WARPS/4, INPUT_PIPE_STAGES = 2;
  using layout = attn_fwd_layout<D, NUM_WORKERS>;
  __device__ static inline void common_setup(common_setup_args<layout> args) {
    args.common.batch = blockIdx.z; args.common.head = blockIdx.y; args.common.seq = blockIdx.x;
    args.num_iters = args.task_iter == 0 ? args.globals.K.rows/layout::kv_tile::rows : -1;
  }
  struct producer {
    __device__ static inline void setup(producer_setup_args<layout> args) {
      warpgroup::producer_registers();
    }
    __device__ static inline void load(producer_load_args<layout> args) {
      if(warpgroup::warpid() == 0) {
        tma::expect(args.inputs_arrived, args.input);
        tma::load_async(args.input.k, args.globals.K, {args.common.batch, args.common.head, args.iter, 0}, args.inputs_arrived);
        tma::load_async(args.input.v, args.globals.V, {args.common.batch, args.common.head, args.iter, 0}, args.inputs_arrived);
      }
      else if(laneid() == 0) arrive(args.inputs_arrived);
    }
  };
  struct consumer {
    __device__ static inline void setup(consumer_setup_args<layout> args) {
      warpgroup::consumer_registers<NUM_WORKERS>();
      if((args.common.seq*NUM_WORKERS + warpgroup::groupid())*layout::qo_tile::rows < args.globals.Q.rows) // out of bounds?
        warpgroup::load(args.scratch.q[warpgroup::groupid()], args.globals.Q,
                       {args.common.batch, args.common.head, args.common.seq*NUM_WORKERS+warpgroup::groupid(), 0});
      zero(args.state.o_reg);
      zero(args.state.norm_vec);
      neg_infty(args.state.max_vec);
      warpgroup::sync(warpgroup::groupid());
    }
    __device__ static inline void compute(consumer_compute_args<layout> args) {
      constexpr float TEMPERATURE_SCALE = (D == 128) ? 0.08838834764f*1.44269504089f : 0.125f*1.44269504089f;
      // A = Q @ K.T
      warpgroup::mm_ABt(args.state.att_block, args.scratch.q[warpgroup::groupid()], args.input.k);
      mul(args.state.max_vec_last_scaled, args.state.max_vec, TEMPERATURE_SCALE);
      warpgroup::mma_async_wait();
\end{lstlisting}
\label{fig:attn-pc-template}
\end{figure}

\begin{figure}[H]
\begin{lstlisting}[basicstyle=\scriptsize\ttfamily]
      // softmax
      row_max(args.state.max_vec, args.state.att_block, args.state.max_vec); // accumulate onto the max_vec
      mul(args.state.max_vec_scaled, args.state.max_vec, TEMPERATURE_SCALE);
      mul(args.state.att_block, args.state.att_block, TEMPERATURE_SCALE);
      sub_row(args.state.att_block, args.state.att_block, args.state.max_vec_scaled);
      exp2(args.state.att_block, args.state.att_block);
      sub(args.state.max_vec_last_scaled, args.state.max_vec_last_scaled, args.state.max_vec_scaled);
      exp2(args.state.max_vec_last_scaled, args.state.max_vec_last_scaled);
      mul(args.state.norm_vec, args.state.norm_vec, args.state.max_vec_last_scaled);
      row_sum(args.state.norm_vec, args.state.att_block, args.state.norm_vec); // accumulate onto the norm_vec
      mul_row(args.state.o_reg, args.state.o_reg, args.state.max_vec_last_scaled); // normalize o_reg before mma
      copy(args.state.att_block_mma, args.state.att_block); // convert to bf16 for mma
      // O += A @ V
      warpgroup::mma_AB(args.state.o_reg, args.state.att_block_mma, args.input.v);
      warpgroup::mma_async_wait();
      if(laneid() == 0) arrive(args.inputs_finished); // done!
    }
    __device__ static inline void finish(consumer_finish_args<layout> args) {
      if((args.common.seq*NUM_WORKERS+warpgroup::groupid())*64 >= args.globals.Q.rows) return; // out of bounds?
      div_row(args.state.o_reg, args.state.o_reg, args.state.norm_vec);
      auto &o_smem = reinterpret_cast<typename layout::qo_tile&>(args.scratch.q[warpgroup::groupid()]);
      warpgroup::store(o_smem, args.state.o_reg);
      warpgroup::sync(warpgroup::groupid());
      if(warpgroup::warpid() == 0)
        tma::store_async(args.globals.O, o_smem, {args.common.batch, args.common.head, args.common.seq*NUM_WORKERS+warpgroup::groupid(), 0});
    }
  };
};
// kernel is kittens::prototype::lcf::kernel<attn_fwd_template<HEAD_DIM>>;
\end{lstlisting}
\label{fig:attn-pc-template}
\caption{A templated non-causal attention kernel for head dims. 64 and 128 that competes with FlashAttention-3.}
\end{figure}
\clearpage

\subsection{Rotary positional encodings}
\label{fig:rotary-pc-template}
This section shows the rotary kernel for head dimension $128$, written in the \shortname{} abstractions.

\begin{figure}[h!]
\begin{lstlisting}[basicstyle=\scriptsize\ttfamily]
using namespace kittens;
using namespace kittens::prototype;
using namespace kittens::prototype::lcsf;
template<int _headdim, int _warps> struct rotary_layout {
  static constexpr int headdim = _headdim, warps = _warps;
  using seq_tile    = st_bf<16, headdim>;
  using seq_global  = gl<bf16, -1, -1, -1, headdim, seq_tile>;
  using rope_global = gl<bf16,  1,  1, -1, headdim/2>;
  struct globals {
    seq_global o, x;
    rope_global sin, cos;
    int batches; // how many batches per block, for sizing grid
  };
  struct input_block    { seq_tile x[warps]; };
  struct output_block   { seq_tile o[warps]; };
  struct producer_state { int active_warps;  };
  struct consumer_state { rt_fl<16, headdim/2> sin, cos; }; // long-resident tiles
};
template<int _headdim> struct rotary_template {
  static constexpr int headdim=_headdim, NUM_CONSUMER_WARPS=8, NUM_BLOCKS=1, OUTPUT_PIPE_STAGES=3, INPUT_PIPE_STAGES=3;
  using layout = rotary_layout<headdim, NUM_CONSUMER_WARPS>;
  __device__ static inline void common_setup(common_setup_args<layout> args) {
    if(args.task_iter == 0) {
      args.num_iters = min(args.globals.batches, (int)(args.globals.x.batch-blockIdx.y*args.globals.batches)) * args.globals.x.depth; // batches*heads handled by block
    }
    else args.num_iters = -1;
  }
  struct producer {
    __device__ static void setup(producer_setup_args<layout> args) {
      warpgroup::producer_registers();
      args.state.active_warps = min((int)NUM_CONSUMER_WARPS,
                                    (int)(args.globals.x.rows/16 - blockIdx.x*NUM_CONSUMER_WARPS));
    }
    __device__ static void load(producer_load_args<layout> args) {
      if(warpgroup::warpid() == args.iter%4) {
          kittens::coord idx = { blockIdx.y*args.globals.batches+args.iter/args.globals.x.depth,
                                 args.iter%args.globals.x.depth,
                                 blockIdx.x*NUM_CONSUMER_WARPS,
                                 0 };
          tma::expect_bytes(args.inputs_arrived, sizeof(layout::seq_tile)*args.state.active_warps);
          for(int i = 0; i < args.state.active_warps; i++) {
              tma::load_async(args.input.x[i], args.globals.x, {idx.b,idx.d,idx.r+i,idx.c}, args.inputs_arrived);
          }
          if(laneid() == 0) arrive(args.inputs_arrived, 3);
          __syncwarp();
      }
    }
    __device__ static void store(producer_store_args<layout> args) {
      if(warpgroup::warpid() == args.iter%4) {
          kittens::coord idx = { blockIdx.y*args.globals.batches+args.iter/args.globals.x.depth,
                                 args.iter%args.globals.x.depth,
                                 blockIdx.x*NUM_CONSUMER_WARPS,
                                 0 };
          for(int i = 0; i < args.state.active_warps; i++) {
              tma::store_async(args.globals.o, args.output.o[i], {idx.b,idx.d,idx.r+i,idx.c});
          }
          tma::store_async_read_wait();
          if(laneid() == 0) arrive(args.outputs_finished, 4);
          __syncwarp();
      }
    }
  };
\end{lstlisting}
\end{figure}

\begin{figure}[H]
\begin{lstlisting}[basicstyle=\scriptsize\ttfamily]
  struct consumer {
    __device__ static void setup(consumer_setup_args<layout> args) {
      warpgroup::consumer_registers<NUM_CONSUMER_WARPS/4>();
      kittens::coord idx = { blockIdx.x*NUM_CONSUMER_WARPS + warpid(), 0 };
      load(args.state.sin, args.globals.sin, idx); // could be better coalesced but doing just once
      load(args.state.cos, args.globals.cos, idx);
    }
    __device__ static void compute(consumer_compute_args<layout> args) {
      rt_fl<16, headdim> x;
      rt_fl<16, headdim/2> x1, x2, temp1, temp2;
      load(x, args.input.x[warpid()]);
      if(laneid() == 0) arrive(args.inputs_finished);
      __syncwarp();
      for(int i = 0; i < headdim/32; i++) {
          #pragma unroll
          for(int j = 0; j < 4; j++) {
              x1.tiles[0][i].data[j] = x.tiles[0][i].data[j];
              x2.tiles[0][i].data[j] = x.tiles[0][i+headdim/32].data[j];
          }
      }
      mul(temp1, x1, args.state.cos);
      mul(temp2, x2, args.state.cos);
      mul(x2, x2, -1.f);
      mul(x1, x1, args.state.sin);
      mul(x2, x2, args.state.sin);
      add(temp1, temp1, x2);
      add(temp2, temp2, x1);
      for(int i = 0; i < headdim/32; i++) {
          #pragma unroll
          for(int j = 0; j < 4; j++) {
              x.tiles[0][i].data[j]            = temp1.tiles[0][i].data[j];
              x.tiles[0][i+headdim/32].data[j] = temp2.tiles[0][i].data[j];
          }
      }
      store(args.output.o[warpid()], x);
      __syncwarp();
      if(laneid() == 0) arrive(args.outputs_arrived);
    }
    __device__ static void finish(consumer_finish_args<layout> args) {
      if(laneid() == 0) arrive(args.finish_finished); // nothing to do here
    }
  };
};
\end{lstlisting}
\label{fig:rotary-pc-template}
\caption{A templated rotary kernel for head dim. 128 that outperforms popular Triton baselines.}
\end{figure}
\clearpage

\section{Shared memory layouts}
\label{app:layouts}
\vspace{-2mm}
To illustrate some of the choices available in shared memory layouts, this appendix outlines six different shared memory layouts for GPU tiles: a naive row-major layout, a padded layout, a simple swizzled layout, and three more specialized swizzled layouts. We are particularly interested in which memory \textit{banks} (numbered from 00 to 31) store each element of the tile; for each layout, we color and label the element of the tile accordingly. We illustrate all layouts using a $32\times64$ 16-bit tile.

\vspace{-3mm}
\subsection{Naive layout}
\begin{figure}[h!]
    \centering
    \includegraphics[width=\textwidth]{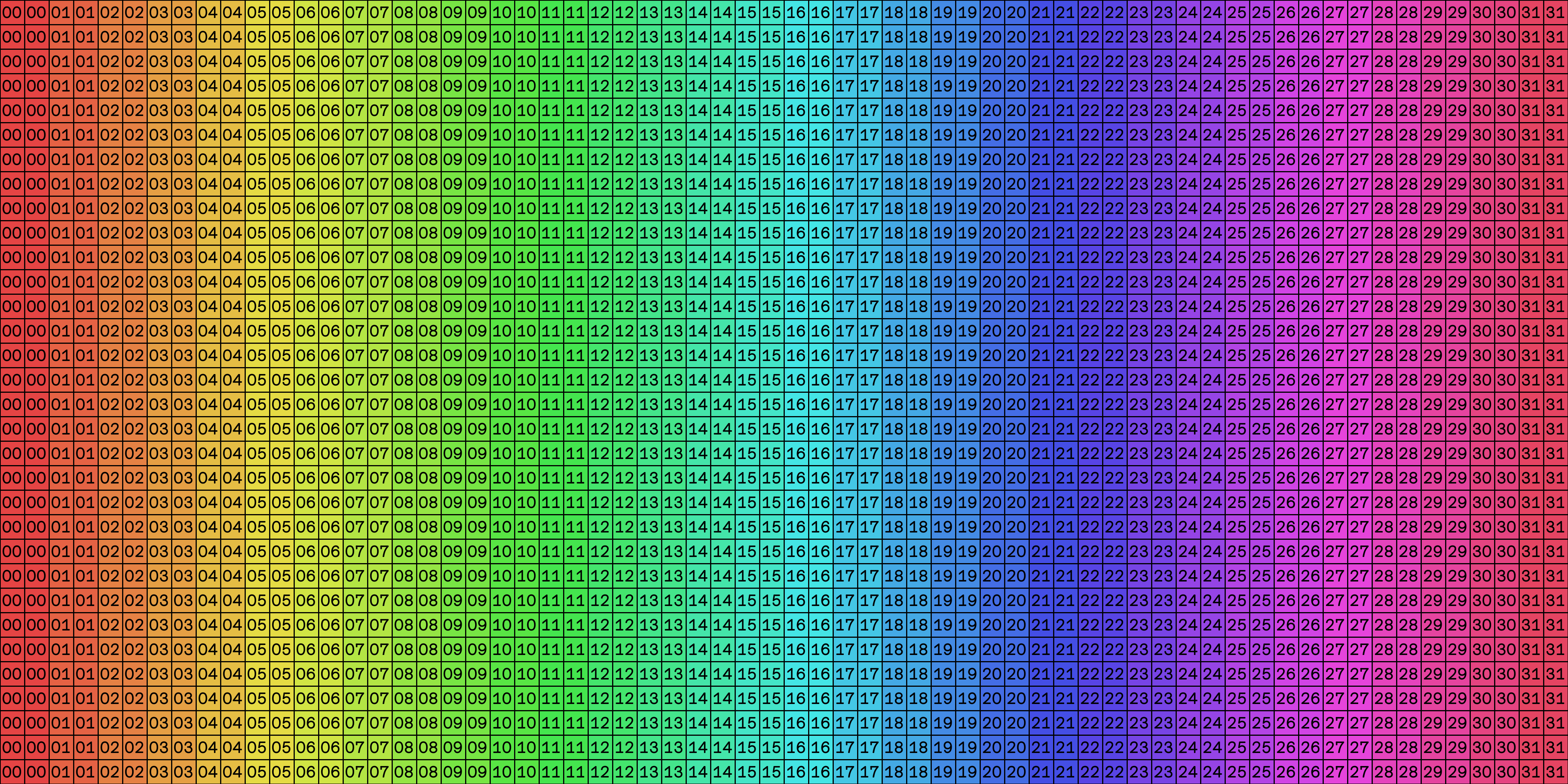}
    \caption{Row-major shared memory layout.}
    \label{fig:naive_layout}
\end{figure}

A row-major layout, illustrated in figure \ref{fig:naive_layout}, is among the simplest layouts. It has the benefit of accurately reflecting tensor layouts in HBM. Furthermore, for access patterns that access row-wise, it has no bank conflicts. But when loading or storing tensor core register layouts, it suffers 8-way bank conflicts, and is thus extremely slow.
\begin{lstlisting}
bf16* naive_layout(bf16 *data, int r, int c) {
    return &data[r * columns + c];
}
\end{lstlisting}

\subsection{Padded layout}
\begin{figure}[H]
    \centering
    \includegraphics[width=\textwidth]{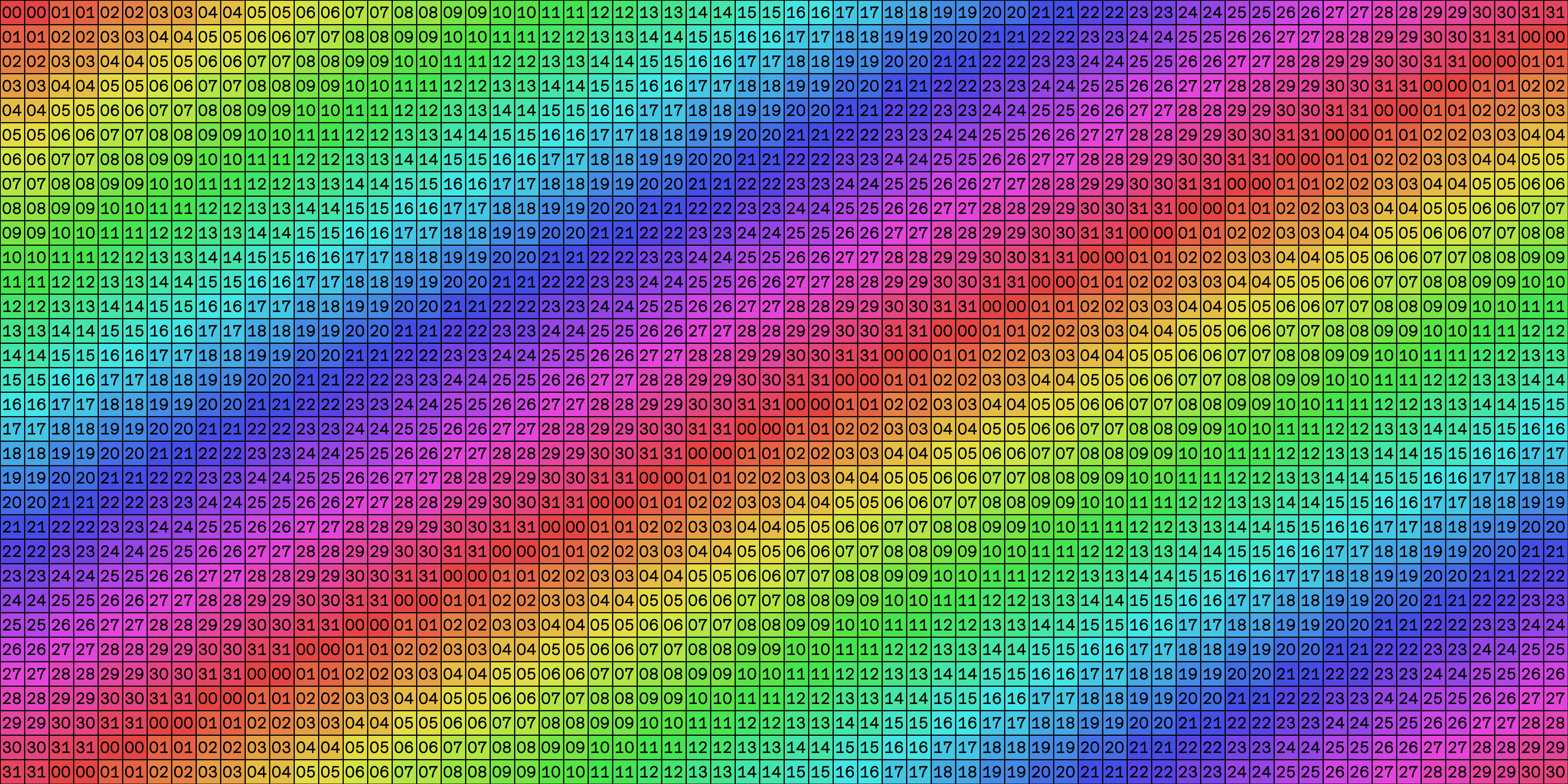}
    \caption{Padded shared memory layout.}
    \label{fig:padded_layout}
\end{figure}

A common solution to these bank conflicts is to ``pad'' each row by one memory bank, thereby introducing an offset to shift consecutive elements of a column into different memory banks. This eliminates bank conflicts, but creates misaligned addresses which interferes with fast instructions that require aligned addresses.
\begin{lstlisting}
bf16* padded_layout(bf16 *data, int r, int c) {
    return &data[r * (columns+1) + c];
}
\end{lstlisting}

\subsection{Naive Swizzled Layout}
\begin{figure}[H]
    \centering
    \includegraphics[width=\textwidth]{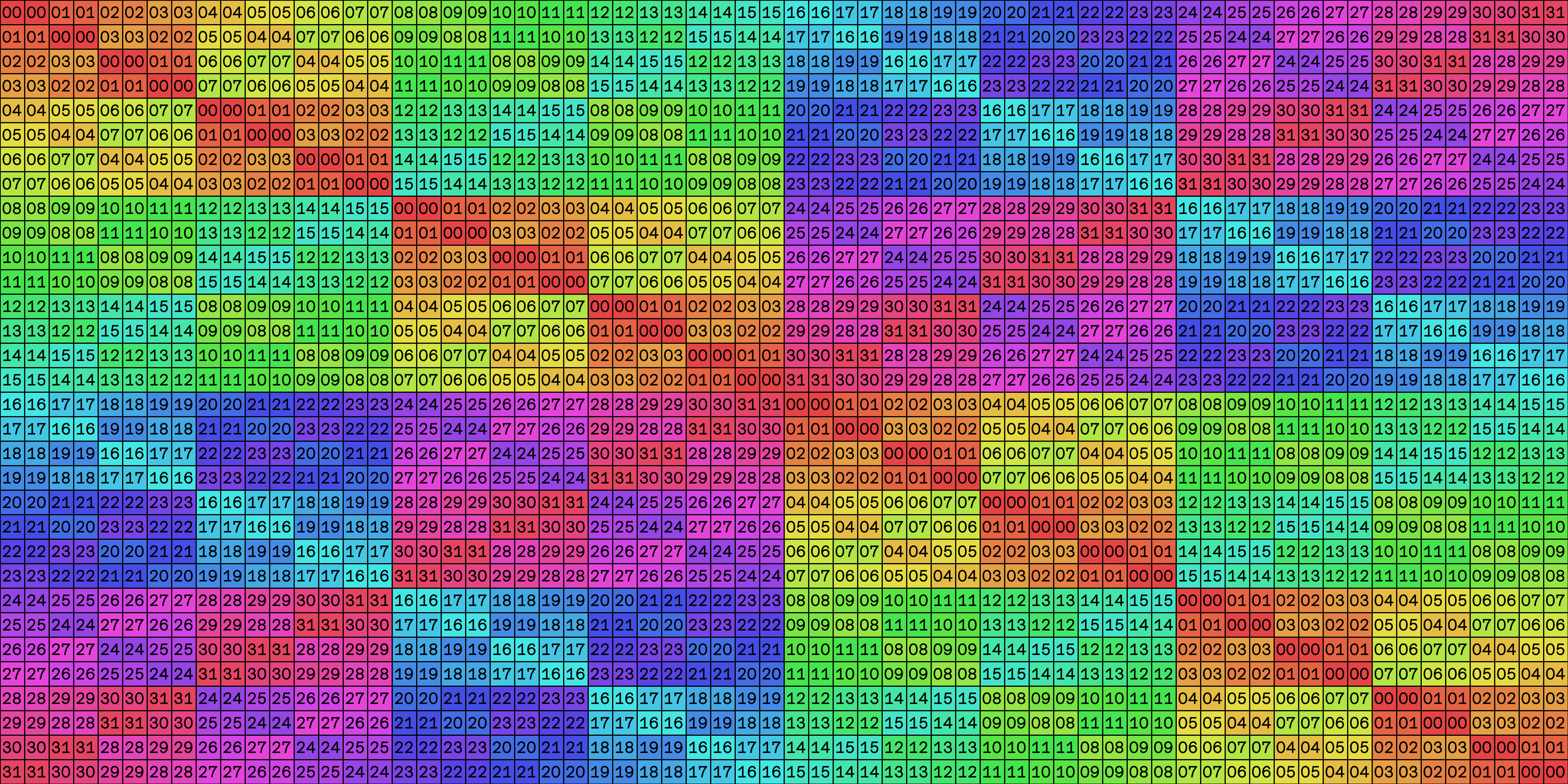}
    \caption{Naive swizzled shared memory layout.}
    \label{fig:naive_swizzled_layout}
\end{figure}

A third option is to ``swizzle'' the memory, in which progressive rows are reshuffled to alter their banking. This layout accomplishes this by xor'ing the index with the row, which eliminates bank conflicts and has aligned addresses. However, this layout lacks hardware support for \codeword{HGMMA} and \codeword{UTMA} instructions, which are particularly important on H100 GPUs for achieving high performance. We illustrate a simple swizzling pattern here:
\begin{lstlisting}
bf16* row_swizzled_layout(bf16 *data, int r, int c) {
    uint64_t addr = (uint64_t)&data[r * columns + c];
    return (bf16*)(addr ^ (r << 2));
}
\end{lstlisting}

\subsection{32 byte swizzling}
\begin{figure}[H]
    \centering
    \includegraphics[width=\textwidth]{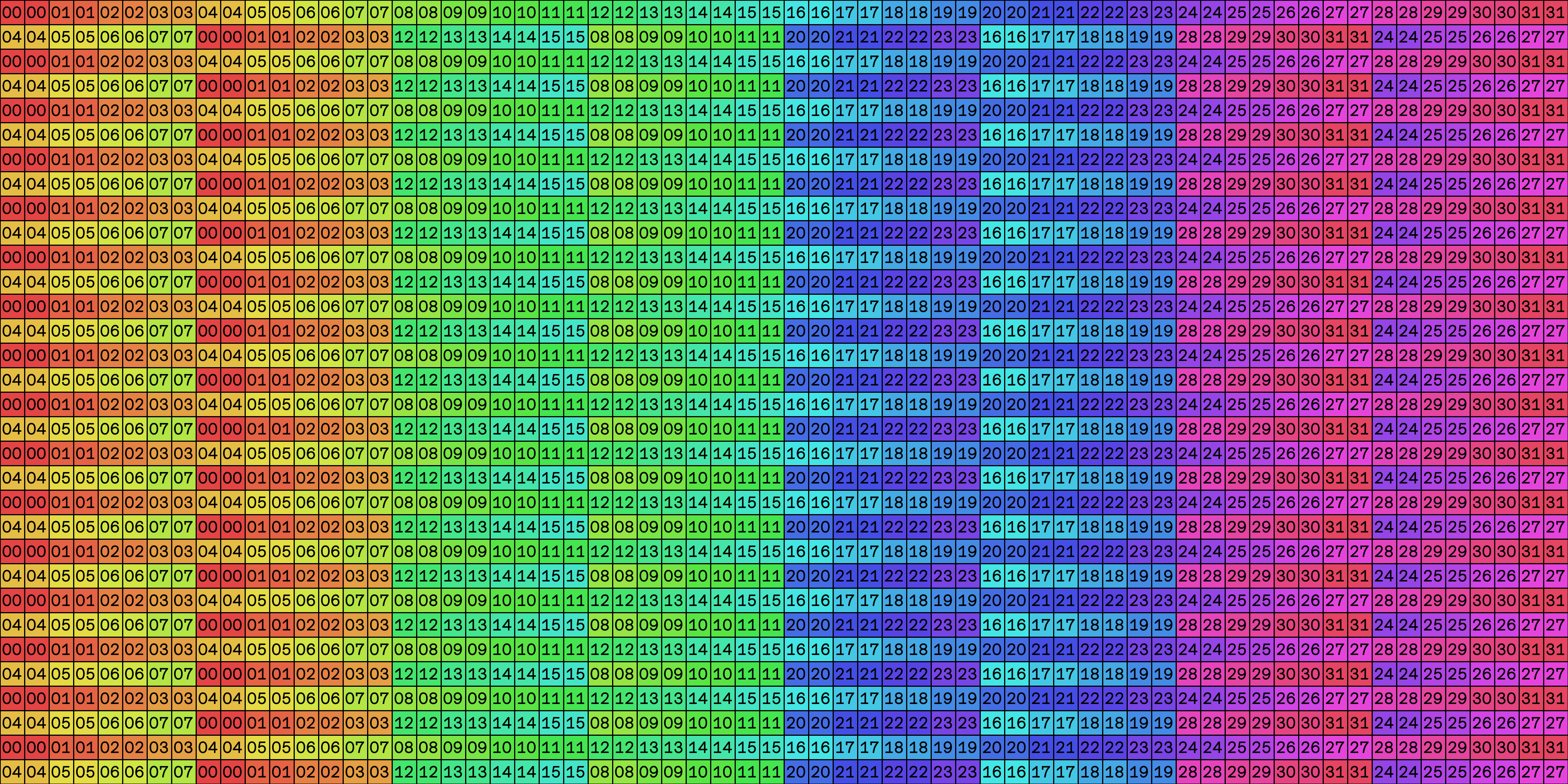}
    \caption{32 byte swizzled shared memory layout.}
    \label{fig:special_swizzled_layout_1}
\end{figure}

32 byte swizzling is the first of a family of layouts (of which we will examine three), where instead of swizzling the index with the row, the memory address is instead swizzled directly with itself. This layout is defined by the following C code:
\begin{lstlisting}
bf16* swizzled_layout_32B(bf16 *data, int r, int c) {
    uint64_t addr = (uint64_t)&data[r * columns + c];
    return (bf16*)(addr ^ (((addr % (32*8)) >> 7) << 4));
}
\end{lstlisting}

This layout suffers from 4-way bank conflicts, but is valid for all tiles whose width is a multiple of 16. However, importantly, it has (as do its siblings below) hardware support from both \codeword{HGMMA} and \codeword{UTMA} instructions.

\subsection{64 byte swizzling}
\begin{figure}[H]
    \centering
    \includegraphics[width=\textwidth]{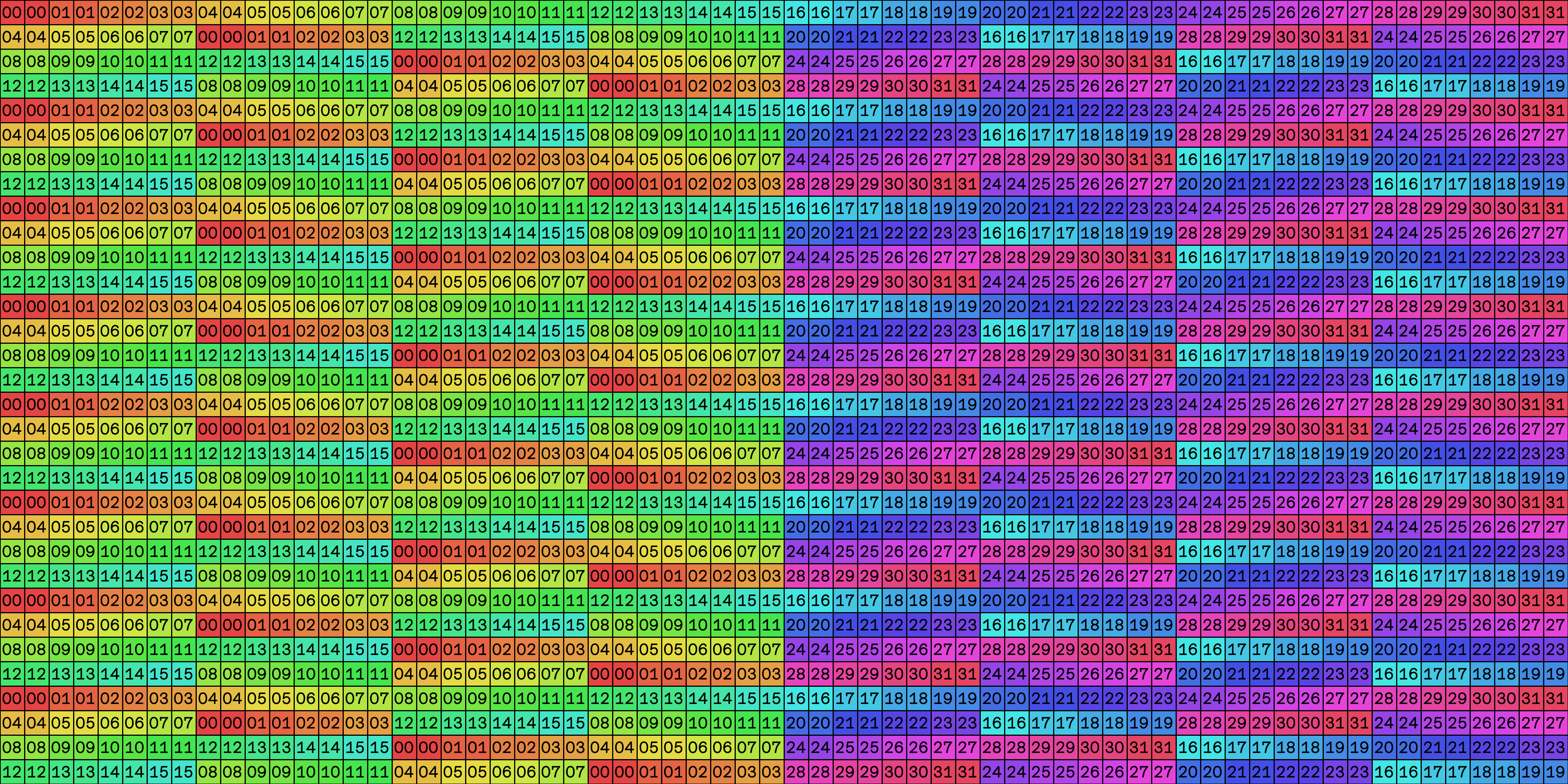}
    \caption{64 byte swizzled shared memory layout.}
    \label{fig:special_swizzled_layout_2}
\end{figure}

64 byte swizzling is a layout similar to 32 byte swizzling with a more aggressive pattern:
\begin{lstlisting}
bf16* swizzled_layout_64B(bf16 *data, int r, int c) {
    uint64_t addr = (uint64_t)&data[r * columns + c];
    return (bf16*)(addr ^ (((addr % (64*8)) >> 7) << 4));
}
\end{lstlisting}

64 byte swizzling suffers from just 2-way bank conflicts, but is only valid for tiles whose width is a multiple of 32 (for half-precision types, or 16 for full-precision).

\subsection{128 byte swizzling.}
\begin{figure}[H]
    \centering
    \includegraphics[width=\textwidth]{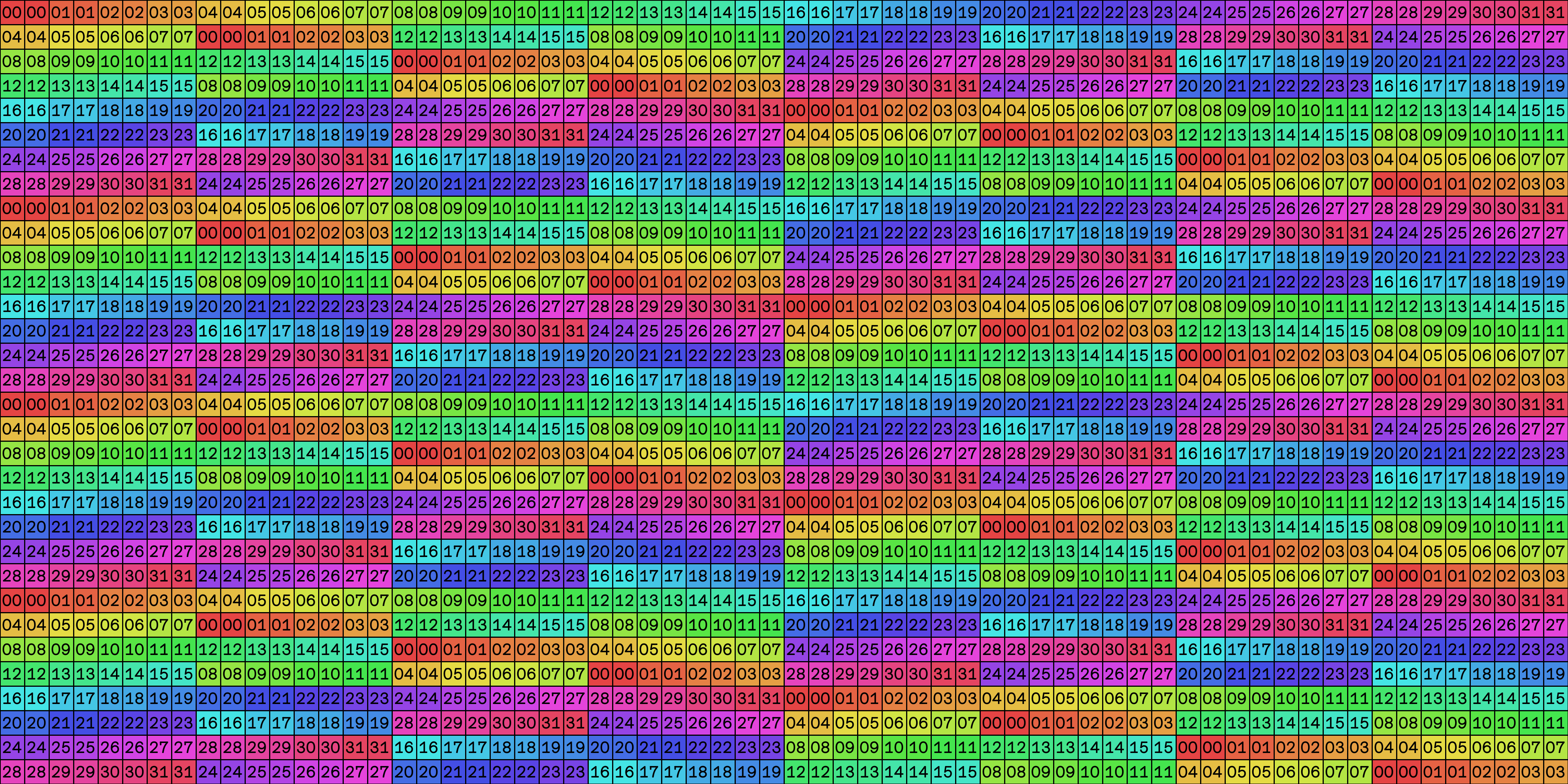}
    \caption{128 byte swizzled shared memory layout.}
    \label{fig:special_swizzled_layout_3}
\end{figure}

128 byte swizzling is a further extension of its kin:
\begin{lstlisting}
bf16* swizzled_layout_128B(bf16 *data, int r, int c) {
    uint64_t addr = (uint64_t)&data[r * columns + c];
    return (bf16*)(addr ^ (((addr % (128*8)) >> 7) << 4));
}
\end{lstlisting}

Finally, 128 byte swizzling has no bank conflicts, but is only valid for half-precision tiles whose width is a multiple of 64.

\subsection{ThunderKittens}

After substantial evaluation of these layouts, we concluded that the three final layouts were the three most important, because \codeword{HGMMA} and \codeword{UTMA} instructions are critical to high performance, and furthermore that they are good enough to yield high performance across many kernels. Correspondingly, depending on the width of the tile at compile time we select the highest level of swizzling possible to minimize bank conflicts.
\end{document}